\documentclass{article}

\usepackage{arxiv}
\usepackage{booktabs}
\usepackage[utf8]{inputenc} 
\usepackage[T1]{fontenc}    
\usepackage[]{hyperref}       
\usepackage{url}            
\usepackage{booktabs}       
\usepackage{amsfonts}       
\usepackage{nicefrac}       
\usepackage{microtype}      
\usepackage{tabularx}
\usepackage{graphicx}
\usepackage{doi}
\usepackage{algorithm}
\usepackage[toc, acronym, automake,nonumberlist]{glossaries}
\setkeys{glslink}{hyper=false}
\usepackage{algpseudocode}
\usepackage{subfig}
\glsdisablehyper
\usepackage{cancel}
\usepackage{array}
\usepackage{rotating}
    
\usepackage{longtable}
\usepackage{lscape}
\usepackage[noabbrev, capitalize]{cleveref}       
\makeglossaries
\newacronym{fox}{FOX}{FOX optimization algorithm}
\newacronym{ifox}{IFOX}{improved FOX}
\newacronym{cec}{CEC}{congress on evolutionary computation}
\newacronym{aco}{ACO}{ant colony optimization}
\newacronym{shoa}{SHOA}{shrike optimization algorithm}
\newacronym{tsp}{TSP}{traveling salesman problem}
\newacronym{leo}{LEO}{lagrange elementary optimization}
\newacronym{ba}{BA}{bat algorithm}
\newacronym{bboa}{BBOA}{brown-bear optimization algorithm}
\newacronym{da}{DA}{dragonfly algorithm}
\newacronym{bda}{BDA}{binary dragonfly algorithm}
\newacronym{fs}{FS}{feature selection}
\newacronym{ga}{GA}{genetic algorithms}
\newacronym{iaga}{IAGA}{improved adaptive genetic algorithm}
\newacronym{gwo}{GWO}{grey wolf optimization}
\newacronym{vw-gwo}{VWGWO}{variable weights grey Wolf optimization}
\newacronym{alo}{ALO}{ant lion optimization}
\newacronym{pso}{PSO}{particle swarm optimization}
\newacronym{ipso}{IPSO}{improved particle swarm optimization}
\newacronym{nro}{NRO}{nuclear reaction optimization}
\newacronym{agv}{AGV}{automated guided vehicles}
\newacronym{fdo}{FDO}{fitness dependent optimizer}
\newacronym{woa}{WOA}{whale optimization algorithm}
\newacronym{iwoa}{IWOA}{improved whale optimization algorithm}
\newacronym{de}{DE}{differential evolution}
\newacronym{fa}{FA}{fox agent}
\newacronym{tid}{TID}{optimization task identifier}
\newacronym{pt}{PT}{processing time}
\newacronym{ana}{ANA}{ant nesting algorithm}
\newacronym{lpb}{LPB}{learner performance-based behavior algorithm}
\newacronym{abc}{ABC}{artificial bee colony algorithms}
\newacronym{sota}{SOTA}{state-of-the-art}

\newacronym{pvd}{PVD}{pressure vessel design}
\newacronym{srp}{SRP}{speed reducer problem}
\newacronym{csd}{CSD}{tension/compression spring design}
\newacronym{pld}{PLD}{piston lever design}
\newacronym{csp}{CSP}{car side impact problem}
\newacronym{bcp}{BCP}{bulk carriers problem}
\newacronym{wbp}{WBP}{welded beam problem}
\newacronym{cbp}{CBP}{cantilever beam problem}
\newacronym{tcp}{TCP}{tubular column problem}
\newacronym{gtp}{GTP}{gear train problem}

\newacronym{vagwo}{VAGWO}{velocity-aided grey wolf optimizer}
\newacronym{ml}{ML}{machine learning}
\newacronym{cpo}{CPO}{crested porcupine optimizer}
\newacronym{coa}{COA}{coati optimization algorithm}
\newacronym{ho}{HO}{hippopotamus optimization algorithm}
\newacronym{lshade}{LSHADE}{Linear population size reduction success-history based adaptive differential evolution}
\newacronym{alshade}{ALSHADE}{adaptive LSHADE}
\newacronym{cma-es}{CMA-ES}{covariance matrix self-adaptation evolutionary algorithm}
\newacronym{ao}{AO}{aquila optimizer}
\newacronym{rsa}{RAS}{reptile search algorithm}
\newacronym{gto}{GTO}{artificial gorilla troops optimizer}
\newacronym{sma}{SMA}{slime mould algorithm}
\newacronym{bmo}{BMO}{barnacles mating optimizer}
\newacronym{bssa}{BSSA}{bear smell search algorithm}
\newacronym{bwoa}{BWOA}{black widow optimization algorithm}
\newacronym{mrfo}{MRFO}{manta ray foraging optimization}
\newacronym{mpa}{MPA}{marine predators algorithm}
\newacronym{ma}{MA}{mayfly algorithm}
\newacronym{rwp}{RWP}{real-world problem}

\usepackage{cancel}
\algnewcommand\algorithmicforeach{\textbf{for each}}
\algdef{S}[FOR]{ForEach}[1]{\algorithmicforeach\ #1\ \algorithmicdo}

\title{An Improved FOX Optimization Algorithm Using Adaptive Exploration and Exploitation for Global Optimization}

\usepackage{authblk}

\newbox{\orcid}\sbox{\orcid}{\includegraphics[scale=0.06]{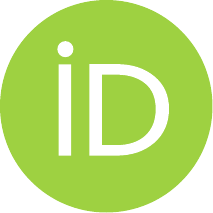}} 
\author[1]{%
    \href{https://orcid.org/0000-0002-6232-3900}{\usebox{\orcid}\hspace{1mm}Mahmood A. Jumaah\thanks{\texttt{cs.22.27@grad.uotechnology.edu.iq}}}%
}
\author[1]{%
    \href{https://orcid.org/0000-0002-7216-4149}{\usebox{\orcid}\hspace{1mm}Yossra H. Ali\thanks{\texttt{Yossra.H.Ali@uotechnology.edu.iq}}}%
}
\author[2]{%
    \href{https://orcid.org/0000-0002-8661-258X}{\usebox{\orcid}\hspace{1mm}Tarik A. Rashid\thanks{\texttt{tarik.ahmed@ukh.edu.krd}}}%
}
\affil[1]{College of Computer Science, University of Technology, Baghdad 10066, Iraq}
\affil[2]{Department of Computer Science and Engineering; AIIC, University of Kurdistan Hewlêr, Erbil 44001, Iraq}

\renewcommand{\shorttitle}{\textit{IFOX}}

\begin{document}
\maketitle
\begin{abstract}
Optimization algorithms are essential for solving many real-world problems. However, challenges such as getting trapped in local minima and effectively balancing exploration and exploitation often limit their performance. This paper introduces an improved variation of the FOX optimization algorithm (FOX), termed Improved FOX (IFOX), incorporating a new adaptive method using a dynamically scaled step-size parameter to balance exploration and exploitation based on the current solution's fitness value. The proposed IFOX also reduces the number of hyperparameters by removing four parameters (C1, C2, a, Mint) and refines the primary equations of FOX. To evaluate its performance, IFOX was tested on 20 classical benchmark functions, 61 benchmark test functions from the congress on evolutionary computation (CEC), and ten real-world problems. The experimental results showed that IFOX achieved a 40\% improvement in overall performance metrics over the original FOX. Additionally, it achieved 880 wins, 228 ties, and 348 losses against 16 optimization algorithms across all involved functions and problems. Furthermore, non-parametric statistical tests, including the Friedman and Wilcoxon signed-rank tests, confirmed its competitiveness against recent and state-of-the-art optimization algorithms, such as LSHADE and NRO, with an average rank of 5.92 among 17 algorithms. These findings highlight the significant potential of IFOX for solving diverse optimization problems, establishing it as a competitive and effective optimization algorithm.
\end{abstract}

\section{Introduction}
Researchers in computer science define optimization as a systematic method to improve systems for problem-solving with maximum performance. It involves identifying the best solutions for a specific problem~\cite{optimizatoin_1}. However, most \glspl{rwp} can be optimized, and with the rapid advancement of computational techniques and increasing computing power, many sectors have experienced significant growth compared to the past~\cite{life_problems_1, benefits_1}. For instance, medicine, engineering, agriculture, commerce, finance, and entertainment~\cite{industry_1,uot_2}. An optimizer is an algorithm designed to solve optimization problems by iteratively searching for the best solution to minimize or maximize an objective function~\cite{optimizer_1, benefits_2}. However, optimizers differ in their problem-solving approach, execution speed, complexity, balance between exploration and exploitation, underlying inspiration, and solution diversity~\cite{optimizer_types_1,life_problems_2}. Improving an optimization algorithm's performance without increasing resource usage means solving problems efficiently with the same processing time and memory consumption to reach the optimal solution, which remains a significant challenge.

This paper concerns the~\gls{fox}, a recent method developed in 2022 by Mohammed and Rashid~\cite{fox}. Although~\gls{fox} has outperformed several optimization algorithms, its broader significance remains limited due to certain limitations. In particular, it has not been evaluated against~\gls{sota} optimizers or tested on a sufficiently wide range of benchmark functions. Additionally, it utilizes a static approach to balance exploration and exploitation by fixing a ratio of 0.5\% for each phase. This does not match the nature of many optimization problems, as some issues need more exploration, and others need more exploitation. For instance, the search space has many local optima in a Rastrigin complex benchmark test function~\cite{rastrigin_figure}. Thus,~\gls{fox} may either converge prematurely or waste resources on unnecessary exploration, as described in the~\Cref{sec:fox_limitations}. Additionally, the mathematical equations used in~\gls{fox} for both the exploration and exploitation phases were acknowledged by its authors as improvable, particularly in terms of jump modeling and prey distance estimation. These limitations reduce its adaptability under challenging problems and thus reduce its overall performance. Therefore, the primary objective of this study is to enhance the performance of~\gls{fox} by addressing its limitations and proposing an improved variation,~\gls{ifox}. We aim to decrease the number of equations and parameters and suggest a completely adaptive method to balance exploration and exploitation. The main contributions of this paper are:

\begin{enumerate}
    \item Replacing constrained equations in~\gls{fox} with more effective alternatives and reducing the number of hyperparameters.
    \item Proposing a new fitness-based adaptive method for better exploration and exploitation balance.
    \item The proposed~\gls{ifox} outperformed the original~\gls{fox} and most of the compared algorithms across various benchmark functions and engineering design problems.
\end{enumerate}

The~\gls{ifox} was evaluated using a wide range of 81 benchmark test functions to ensure rigorous performance evaluations. These included 20 classical functions (ten unimodal and ten multimodal) and 61~\gls{cec}-based functions from~\gls{cec} 2017,~\gls{cec} 2019,~\gls{cec} 2021, and~\gls{cec} 2022, which are widely used in optimization algorithms assessments. Additionally, ten \glspl{rwp} were employed to demonstrate the applicability of~\gls{ifox} in real-world, as explained in~\Cref{sec:real_world_problems}. However, the experimental results show that the proposed~\gls{ifox} outperforms the basic~\gls{fox} while achieving performance comparable to various recent and~\gls{sota} optimization algorithms.

The rest of this paper is organized as follows: The~\Cref{sec:literature_review} provides a literature review of existing improved optimization algorithms and their applications, in addition to our previous optimization algorithms. The~\Cref{sec:methodology} describes the methods used in this paper to develop the~\gls{ifox}. The~\Cref{sec:results} lists the results obtained from the experiments. The~\Cref{sec:discussion} discusses, analyzes, and compares the findings. Finally, The~\Cref{sec:conclusion} concludes and summarizes the key contributions and limitations and suggests future work directions.
\section{Literature review\label{sec:literature_review}}
Optimization algorithms are advancing rapidly, making significant progress in many industries and fields, especially in engineering and computer science~\cite{related_optimizatoin_in_cs}. New algorithms are frequently proposed and subsequently improved by researchers to enhance their global problem-solving capabilities or to adapt them to specific types of problems~\cite{related_optimizatoin_problems_1, related_optimizatoin_problems_2, related_optimizatoin_problems_3}. Given the continuous emergence of new challenges, the development and enhancement of optimization algorithms remain a highly relevant and hot-topic area of research~\cite{related_hot_topic_area, uot_1}. The following paragraphs present optimization algorithms categorized by their inspiration source, followed by recent and~\gls{sota} algorithms.

Optimization algorithms can be classified into many categories: swarm-based algorithms, evolutionary algorithms, physics or chemistry-based algorithms, or in hybrid form. However, algorithms that take inspiration from the collective actions of species like ants, fish, bees, wasps, and birds are known as swarm intelligence algorithms~\cite{swarm_intelligence}. This concept comes from the natural group behavior of these species as they search for food. A key feature of swarm-based algorithms is that they rely on simple agents that are not complex. These agents work together through indirect communication and move within the decision space to find better solutions~\cite{swarm_intelligence1}.
For instance,~\gls{shoa} is a popular optimization method that takes inspiration from how shrike birds migrate, keep alive and reproduce. Using this algorithm, both stages of exploration and exploitation are structured, just like birds feed and take care of their young until they are able to look after themselves. A mathematical expression was developed for the algorithm and it was also tested on 41 different problems in optimization. This meant using 19 benchmark test functions, 10 CEC-2019 and 12 CEC-2022 and also four engineering design problems with constraints and no constraints. According to the research,~\gls{shoa} shows excellent performance when handling tasks that need to optimize multiple variables together~\cite{related_SHOA}.

Similarly,~\gls{gwo} is commonly used algorithm that take his inspiration from wolves foraging behavior. Various studies focused on improving the performance of the basic~\gls{gwo}. For instance, a~\gls{vagwo} was proposed to enhance global search and avoid rash convergence~\cite{improved_gwo_velocity_based}. Another improvement applied an exponential decay approach to better control the transition from exploration to exploitation throughout the optimization process by introducing a differential perturbation using three omega wolves and applying unified 
$A$ and $C$ parameters to promote exploration during exploitation~\cite{enhanced_Search_gwo}. Furthermore, the~\gls{gwo} has been modified to solve feature selection in high-dimensional data by integrating ReliefF and Copula entropy during initialization. It utilizes a competitive guidance strategy for flexible search and a differential evolution-based method to enhance leader positioning and avoid local optima~\cite{gwo_fs}. Additionally, an improved version of~\gls{gwo} called~\gls{vw-gwo} has been proposed, utilizes social hierarchies to avoid local optima and performs well in high-dimensional scenarios. It used a modified control parameter equation to reduce the local optima trap by proposing that in the standard~\gls{gwo}, all dominant wolves influence the search equally, but this contradicts the social hierarchy of grey wolves. Since the alpha is the true leader, its position should have the highest influence, followed by the beta and the delta. Early in the search, the alpha's position should either be the sole reference or weighted more heavily to reflect its dominant role~\cite{related_vw-gwo}. The results of~\gls{vw-gwo} indicate that it surpasses basic~\gls{gwo},~\gls{alo} and~\gls{pso} across several benchmark functions, which makes it a suitable choice for comparison and analysis purposes in this study. 

Bio-inspired optimization algorithms, inspired by biological phenomena such as evolution and animal behavior, are widely used across many disciplines~\cite{related_power_optimization}. For example,~\Gls{bboa} uses brown bear communication capabilities to balance search exploration and exploitation through its sniffing mechanisms and footprint markings, which prove superiority in minimizing power system expenses~\cite{related_bboa}. An additional bio-inspired algorithm, named~\gls{da}, inspired by the hunting and migration behaviors of dragonflies, simulates two key behaviors: static hunting, where dragonflies create a small, static swarm to encircle prey, and dynamic migration, where they form more prominent groups to travel long distances~\cite{related_da}. The~\gls{da} has been extended to its binary form~\gls{bda} to handle binary optimization tasks such as~\gls{fs}. It updates the main coefficients—separation, alignment, cohesion, food attraction, and enemy distraction—using random values, aiming to imitate the natural movement and interactions of dragonflies in a binary space. Furthermore, experimental results demonstrate that the~\gls{bda} achieves superior performance, selects fewer yet more relevant features, and attains better objective function values across 18 well-known benchmark datasets~\cite{related_bda}. Moreover, recent study introduces an~\gls{ipso} for~\gls{agv} path planning problem in a one-line production. This improved algorithm incorporates a novel coding method, crossover operation, and mutation mechanism to refine solution quality. The experiments showed the superior performance of~\gls{ipso} over the conventional optimization algorithms, underlining its efficacy in minimizing transportation time and avoiding local optima~\cite{related_ipso}.

Evolutionary algorithms are stochastic population-based metaheuristics widely used to solve complex and various issues in domains such as optimization, system modeling, and engineering design. This advancement made evolutionary computation an independent research domain. Hybrid evolutionary approaches combine various optimization algorithms to enhance performance. For instance, in a recent study, researchers investigated the limitations of traditional~\gls{ga}, which, despite being powerful tools for global optimization based on evolution theory, traditional~\gls{ga} tend to have suboptimal convergence speed and efficiency. However, to tackle this issue, the study suggests an~\gls{iaga} that improves the crossover and mutation probabilities adaptively based on the fitness value of individuals in the group. The results of evaluating~\gls{iaga} on~\gls{tsp} problems achieves faster convergence and improved efficiency~\cite{related_iaga}. Moreover~\gls{woa} is improved through hybridization with~\gls{de} to proposes the~\gls{iwoa} by combining~\gls{woa}'s exploitation with~\gls{de}'s exploration, addressing~\gls{woa}'s early convergence issues. Additionally, an extended version,~\gls{iwoa}+, incorporates re-initialization and adaptive parameter control. Comparative experiments demonstrate that both~\gls{iwoa} and~\gls{iwoa}+ outperform other algorithms regarding solution quality and convergence across 25 benchmark test functions~\cite{related_iwoa}.

The~\gls{sota} algorithms represent current benchmarks in optimization research. Among~\gls{sota} evolutionary algorithms,~\gls{lshade} enhances~\gls{de} by incorporating a linear population size reduction strategy and adaptive parameter control mechanisms. This approach significantly boosts convergence speed and robustness, consistently achieving superior results across complex benchmark suites like~\gls{cec} 2014 and~\gls{cec} 2017~\cite{lshade}. Furthermore, the~\gls{alshade} algorithm extends~\gls{lshade} by adaptively adjusting strategies and parameters during optimization runs. It has shown significant performance improvements in solving challenging optimization problems from~\gls{cec} 2017 and~\gls{cec} 2020 benchmark test functions~\cite{alshade}. Additionally, the~\gls{cma-es} adapts the covariance matrix of the mutation distribution dynamically, allowing efficient learning of the problem landscape structure. It is widely recognized for its robustness and reliability in solving complex global optimization problems, demonstrating superior results in various benchmark test suites, including~\gls{cec} competitions and numerous real-world applications~\cite{cma-es}. 

Several recent optimization algorithms further advance the optimization landscape, highlighting the continuous innovation in metaheuristic design. These include~\gls{cpo} is a bio-inspired algorithm that operates on porcupines' defensive patterns and achieves population reduction cycling for exploration and exploitation control~\cite{cpo}; the~\gls{coa} is inspired by two distinct natural behaviors of coatis~\cite{coa}; the~\gls{ho} introduces a novel three-phase behavioral model inspired by hippopotamus activities~\cite{HO}; the~\gls{nro} is physics-inspired algorithm of nuclear fission and fusion phases~\cite{related_nro}; the~\gls{leo} is a self-adaptive evolutionary method motivated by the precision of vaccinations using the albumin quotient of human blood. In~\gls{leo}, intelligent search agents are developed based on their fitness values following a genetic crossover process. These gene-based mechanisms guide the agents through both the exploration and exploitation processes. The core idea and motivation behind~\gls{leo} are introduced in~~\cite{leo2023}; the~\gls{ao}, inspired by eagle hunting strategies~\cite{ao_optimizer}; the~\gls{rsa}, based on reptilian search behaviors~\cite{rsa_optimizer}; the~\gls{gto}, modeling social dynamics of gorilla troops~\cite{gto_optimizer}; the~\gls{sma}, derived from the foraging patterns of slime moulds~\cite{sma_optimizer}; the~\gls{bmo}, reflecting barnacle mating behavior~\cite{bmo_optimizer}; the~\gls{bssa}, inspired by bear olfactory tracking~\cite{bssa_optimizer}; the~\gls{bwoa}, simulating black widow spider mating and cannibalism~\cite{bwoa_optimizer}; the~\gls{mrfo}, based on manta ray foraging~\cite{mrfo_optimizer}; the~\gls{mpa}, utilizing predator motion patterns like lévy and brownian movements~\cite{mpa_optimizer}; and the~\gls{ma}, driven by the mating behavior of mayflies~\cite{ma_optimizer}. These optimization algorithms are emphasized in recent surveys~\cite{survey_optimization_2023, survey_optimization_2023_1, survey_optimization_2025}, and demonstrate strong potential in solving complex optimization challenges.

Table~\ref{tab:literature_review_summary} presents the~\gls{sota} and recent optimization algorithms reviewed, including the name, type, inspiration or main features, and evaluation method. The literature review showed that many algorithms have been improved from their basic form to overcome challenges and limitations and achieve better performance. Common limitations include tuning parameters, escaping local optima, and balancing exploration and exploitation. Based on these observations and an analysis of the limitations of the~\gls{fox}, as will be explained in the~\Cref{sec:fox_limitations}, this study aims to address the existing gaps by proposing an improved and more reliable variation of~\gls{fox}, capable of enhancing search efficiency by proposing a new method for balancing exploration and exploitation.

\begin{table}[htb]
        \centering
        \caption{Summary of recent and state-of-the-art optimization algorithms}
        \label{tab:literature_review_summary}
        \begin{tabularx}{\linewidth}{|l|l|X|l|l|}
            \hline
            \textbf{Algorithm} & \textbf{Type} & \textbf{Main Feature / Inspiration}           & \textbf{Evaluated On} & \textbf{Ref.}                      \\
            \hline
            ACO                & Swarm-based   & Classic pheromone trail search                & TSP                   & \cite{related_aco}                 \\ \hline
            ALSHADE            & Evolutionary  & Self-tuned LSHADE                             & CEC 2017, CEC 2020    & \cite{alshade}                     \\ \hline
            AO                 & Bio-inspired  & Eagle hunting strategies                      & CEC 2021              & \cite{ao_optimizer}                \\ \hline
            BDA                & Bio-inspired  & Binary dragonfly for feature selection        & 18 FS datasets        & \cite{related_bda}                 \\ \hline
            BBOA               & Bio-inspired  & Bear olfactory tracking                       & Power systems         & \cite{related_bboa}                \\ \hline
            BMO                & Bio-inspired  & Barnacle mating process                       & CEC 2022              & \cite{bmo_optimizer}               \\ \hline
            BSSA               & Bio-inspired  & Bear smell tracking                           & CEC 2022              & \cite{bssa_optimizer}              \\ \hline
            BWOA               & Bio-inspired  & Widow spider mating/cannibalism               & CEC 2020              & \cite{bwoa_optimizer}              \\ \hline
            CMA-ES             & Evolutionary  & Covariance-based mutation                     & CEC, real-world       & \cite{cma-es}                      \\ \hline
            COA                & Bio-inspired  & Coati foraging and climbing                   & CEC 2020              & \cite{coa}                         \\ \hline
            CPO                & Bio-inspired  & Porcupine defensive behavior                  & CEC 2021              & \cite{cpo}                         \\ \hline
            DA                 & Bio-inspired  & Swarm hunting and migration                   & Standard functions    & \cite{related_da}                  \\ \hline
            FOX                & Bio-inspired  & Distance-controlled exploration               & CEC 2019              & \cite{fox}                         \\ \hline
            GA                 & Evolutionary  & Canonical genetic operators                   & General               & \cite{related_iaga}                \\ \hline
            GTO                & Bio-inspired  & Gorilla troop behavior                        & CEC 2021              & \cite{gto_optimizer}               \\ \hline
            GWO-FS             & Swarm-based   & Feature selection with entropy initialization & High-dim FS tasks     & \cite{gwo_fs}                      \\ \hline
            HO                 & Bio-inspired  & Hippopotamus activity modeling                & CEC 2022              & \cite{HO}                          \\ \hline
            IAGA               & Evolutionary  & Fitness-based adaptive GA                     & TSP                   & \cite{related_iaga}                \\ \hline
            SHOA            & Swarm-based   & mimics shrike bird behavior for efficient optimization        & CEC 2019, CEC 2022  & \cite{related_SHOA}             \\ \hline
            IPSO               & Bio-inspired  & Enhanced PSO for AGV routing                  & Real-world AGV        & \cite{related_ipso}                \\ \hline
            IWOA               & Hybrid        & DE-exploration + WOA exploitation             & CEC 2017, CEC 2020    & \cite{related_iwoa}                \\ \hline
            IWOA+              & Hybrid        & IWOA with reinitialization                    & CEC 2020              & \cite{related_iwoa}                \\ \hline
            LEO                & Bio-inspired  & Learning via albumin quotient                 & CEC 2023              & \cite{leo2023}                     \\ \hline
            LSHADE             & Evolutionary  & Adaptive DE with shrinking population         & CEC 2014, CEC 2017    & \cite{lshade}                      \\ \hline
            MA                 & Bio-inspired  & Mayfly drift and mating                       & CEC 2021              & \cite{ma_optimizer}                \\ \hline
            MPA                & Bio-inspired  & Predator Lévy and Brownian paths              & CEC 2021              & \cite{mpa_optimizer}               \\ \hline
            MRFO               & Bio-inspired  & Manta ray motion modes                        & CEC 2020              & \cite{mrfo_optimizer}              \\ \hline
            NRO                & Physics-based & Nuclear reaction phases                       & CEC 2020              & \cite{related_nro}                 \\ \hline
            RSA                & Bio-inspired  & Reptile-based adaptive movement               & CEC 2022              & \cite{rsa_optimizer}               \\ \hline
            SMA                & Bio-inspired  & Slime mould foraging                          & CEC 2020              & \cite{sma_optimizer}               \\ \hline
            VAGWO              & Swarm-based   & Velocity-guided wolf search                   & CEC 2019              & \cite{improved_gwo_velocity_based} \\ \hline
            VWGWO              & Swarm-based   & Weighted leadership in wolf hierarchy         & CEC 2017, CEC 2020    & \cite{related_vw-gwo}              \\ \hline
        \end{tabularx}
\end{table}

\section{Methodology\label{sec:methodology}}
This section presents the methods for improving the~\gls{fox}. It begins by describing the benchmark test functions and \glspl{rwp} used to evaluate the performance of the proposed algorithm. The subsequent section analyzes the basic~\gls{fox} and its limitations, followed by a detailed description of the proposed improvements introduced in~\gls{ifox}, along with supporting algorithms and flowcharts.

\subsection{Benchmark test functions}
This section presents the benchmark test function that includes classical and~\gls{cec}-based from the~\gls{cec} 2017, 2019, 2021, and 2022.  However, a specific~\gls{tid} will used instead of the full name of the function to increase the clarity of results and discussion presentation. Classical functions will be denoted as \textit{CL}, and~\gls{cec}-based functions are refered as follows: \textit{C17} for~\gls{cec} 2017, \textit{C19} for~\gls{cec} 2019, \textit{C21} for~\gls{cec} 2021, and \textit{C22} for~\gls{cec} 2022.

Classical benchmark test functions are widely used in computational optimization to evaluate and compare the performance of algorithms. These functions serve as standardized tests due to their diverse mathematical properties, such as modality, separability, and dimensionality. Unimodal functions, characterized by a single global minimum, assess the exploitation capability of an algorithm—its ability to converge to the optimum in smooth landscapes. In contrast, multimodal functions contain multiple local minima and evaluate the exploration capability of algorithms, such as avoiding early convergence and determining the global optimum in complex search spaces. In this paper, 20 benchmark functions were selected, including ten unimodal and ten multimodal functions. These functions were chosen for their varied mathematical properties and recognition in the optimization literature~\cite{why_needs_test_optimizer,sphare_test_function,classical_optimization_functions}. On the other hand, The~\gls{cec} is considered one of the most significant conferences within evolutionary computation, managed by IEEE, which presents complex benchmark test functions. These functions are more complicated than classical ones created to closely mimic real-world optimization challenges. They incorporate a variety of transformations and hybridizations to increase the complexity. Specifically, these functions are constructed using the following features.~\textit{Shifting} involves moving the global optimum to another point than the origin, thereby allowing algorithms to avoid any bias that centers around the origin. In addition,~\textit{rotation} refers to the coordinate system of the function being rotated to introduce dependencies among the decision variables, making the problem non-separable and more challenging. Moreover,~\textit{hybridization} combines different classical functions through weighted sums or division of decision space to form unified functions. The implementation of hybridization introduces various landscape features alongside multiple modes throughout other regions. Furthermore,~\textit{composition} combines two or more transformed functions into one entity, creating an unstable search landscape that misguides the decision-making process. Finally,~\textit{dimension} and range variability mean the test functions cover a wide range of dimensions and decision variable bounds to assess scalability and robustness~\cite{benchmark_test_functions}. The Table~\ref{tab:benchmark_test_functions} provides 81 classical and~\gls{cec}-based functions as follows: 20 classical~\cite{classical_optimization_functions}, 29 from~\gls{cec} 2017~\cite{cec2017}, ten from~\gls{cec} 2019 \cite{cec2019}, ten from~\gls{cec} 2021 \cite{cec2021}, and 12 from the~\gls{cec} 2022 \cite{cec2022}. The implementations of the classical and~\gls{cec}-based benchmark test functions used in this study are available at~\url{https://github.com/thieu1995/opfunu}.
\begin{longtable}{|l|l|l|l|l|l|}
    \caption{Summary of classical and~\gls{cec}-based benchmark test functions used in this study \label{tab:benchmark_test_functions}}      \\
                                                                                  
    \hline
    \textbf{TID} & \textbf{Name}                                                & \textbf{Type} & \textbf{Range}  & \textbf{Dim} & $f_{min}$ \\
    \hline
    \endfirsthead 
    \multicolumn{6}{@{}l}{\textit{\ldots continued from previous page}}                                                                      \\
                                                                                   
    \hline
    \textbf{TID} & \textbf{Name}                                                & \textbf{Type} & \textbf{Range}  & \textbf{Dim} & $f_{min}$ \\
    \hline
    \endhead
    
    \multicolumn{6}{r@{}}{\textit{continued on next page \ldots}}                                                                            \\
    \endfoot    
    \endlastfoot     
    CL1          & Ackley 02 function                                           & Unimodal      & [-32, 32]       & 2            & -200      \\ \hline
    CL2          & Beale function                                               & Unimodal      & [-4.5, 4.5]     & 2            & 0         \\ \hline
    CL3          & Booth function                                               & Unimodal      & [-10, 10]       & 2            & 0         \\ \hline
    CL4          & Chen Bird function                                           & Unimodal      & [-500, 500]     & 2            & -2000     \\ \hline
    CL5          & Chung Reynolds function                                      & Unimodal      & [-100, 100]     & 2            & 0         \\ \hline
    CL6          & Cube function                                                & Unimodal      & [-10, 10]       & 2            & 0         \\ \hline
    CL7          & Dixon \& Price function                                      & Unimodal      & [-10, 10]       & 2            & 0         \\ \hline
    CL8          & Brent function                                               & Unimodal      & [-10, 10]       & 2            & 0         \\ \hline
    CL9          & Leon function                                                & Unimodal      & [-1.2, 1.2]     & 2            & 0         \\ \hline
    CL10         & Matyas function                                              & Unimodal      & [-10, 10]       & 2            & 0         \\ \hline
    CL11         & Ackley 01 function                                           & Multimodal    & [-35, 35]       & 2            & 0         \\ \hline
    CL12         & Adjiman function                                             & Multimodal    & [-1, 2]         & 2            & -2.022    \\ \hline
    CL13         & Alpine01 function                                            & Multimodal    & [-10, 10]       & 2            & 0         \\ \hline
    CL14         & Bird function                                                & Multimodal    & [-6.2, 6.2]     & 2            & -106.7    \\ \hline
    CL15         & Camel 3-Hump function                                        & Multimodal    & [-5, 5]         & 2            & 0         \\ \hline
    CL16         & Dolan function                                               & Multimodal    & [-100, 100]     & 5            & 0         \\ \hline
    CL17         & Exponential function                                         & Multimodal    & [-1, 1]         & 2            & -1        \\ \hline
    CL18         & Griewank function                                            & Multimodal    & [-100, 100]     & 2            & 0         \\ \hline
    CL19         & Hartman 3 function                                           & Multimodal    & [0, 1]          & 3            & -3.863    \\ \hline
    CL20         & McCormick                                                    & Multimodal    & [-1.5, 4]       & 2            & -1.913    \\ \hline
    C17F1        & Shifted and Rotated Bent Cigar function                      & Unimodal      & [-100.0, 100.0] & 30           & 100       \\ \hline
    C17F2        & Shifted and Rotated Zakharov function                        & Unimodal      & [-100, 100]     & 30           & 200       \\ \hline
    C17F3        & Shifted and Rotated Rosenbrock's function                    & Multimodal    & [-100, 100]     & 30           & 300       \\ \hline
    C17F4        & Shifted and Rotated Rastrigin's function                     & Multimodal    & [-100, 100]     & 30           & 400       \\ \hline
    C17F5        & Shifted and Rotated Schaffer's F7 function                   & Multimodal    & [-100, 100]     & 30           & 500       \\ \hline
    C17F6        & Shifted and Rotated Lunacek Bi-Rastrigin's function          & Multimodal    & [-100, 100]     & 30           & 600       \\ \hline
    C17F7        & Shifted and Rotated Non-Continuous Rastrigin's function      & Multimodal    & [-100, 100]     & 30           & 700       \\ \hline
    C17F8        & Shifted and Rotated Levy function                            & Multimodal    & [-100, 100]     & 30           & 800       \\ \hline
    C17F9        & Shifted and Rotated Schwefel's function                      & Multimodal    & [-100, 100]     & 30           & 900       \\ \hline
    C17F10       & Hybrid function 1                                            & Hybrid        & [-100, 100]     & 30           & 1000      \\ \hline
    C17F11       & Hybrid function 2                                            & Hybrid        & [-100, 100]     & 30           & 1100      \\ \hline
    C17F12       & Hybrid function 3                                            & Hybrid        & [-100, 100]     & 30           & 1200      \\ \hline
    C17F13       & Hybrid function 4                                            & Hybrid        & [-100, 100]     & 30           & 1300      \\ \hline
    C17F14       & Hybrid function 5                                            & Hybrid        & [-100, 100]     & 30           & 1400      \\ \hline
    C17F15       & Hybrid function 6                                            & Hybrid        & [-100, 100]     & 30           & 1500      \\ \hline
    C17F16       & Hybrid function 7                                            & Hybrid        & [-100, 100]     & 30           & 1600      \\ \hline
    C17F17       & Hybrid function 8                                            & Hybrid        & [-100, 100]     & 30           & 1700      \\ \hline
    C17F18       & Hybrid function 9                                            & Hybrid        & [-100, 100]     & 30           & 1800      \\ \hline
    C17F19       & Hybrid function 10                                           & Hybrid        & [-100, 100]     & 30           & 1900      \\ \hline
    C17F20       & Composition function 1                                       & Composition   & [-100, 100]     & 30           & 2000      \\ \hline
    C17F21       & Composition function 2                                       & Composition   & [-100, 100]     & 30           & 2100      \\ \hline
    C17F22       & Composition function 3                                       & Composition   & [-100, 100]     & 30           & 2200      \\ \hline
    C17F23       & Composition function 4                                       & Composition   & [-100, 100]     & 30           & 2300      \\ \hline
    C17F24       & Composition function 5                                       & Composition   & [-100, 100]     & 30           & 2400      \\ \hline
    C17F25       & Composition function 6                                       & Composition   & [-100, 100]     & 30           & 2500      \\ \hline
    C17F26       & Composition function 7                                       & Composition   & [-100, 100]     & 30           & 2600      \\ \hline
    C17F27       & Composition function 8                                       & Composition   & [-100, 100]     & 30           & 2700      \\ \hline
    C17F28       & Composition function 9                                       & Composition   & [-100, 100]     & 30           & 2800      \\ \hline
    C17F29       & Composition function 10                                      & Composition   & [-100, 100]     & 30           & 2900      \\ \hline
    C19F1        & Storn's Chebyshev Polynomial Fitting Problem                 & Multimodal    & [-8192, 8192]   & 9            & 1         \\ \hline
    C19F2        & Inverse Hilbert Matrix                                       & Multimodal    & [-16384, 16384] & 16           & 1         \\ \hline
    C19F3        & Lennard-Jones Minimum Energy Cluster                         & Multimodal    & [-4, 4]         & 18           & 1         \\ \hline
    C19F4        & Shifted and rotated Rastrigin's function                     & Multimodal    & [-100, 100]     & 10           & 1         \\ \hline
    C19F5        & Shifted and rotated Griewank's function                      & Multimodal    & [-100, 100]     & 10           & 1         \\ \hline
    C19F6        & Shifted and rotated Weierstrass function                     & Multimodal    & [-100, 100]     & 10           & 1         \\ \hline
    C19F7        & Shifted and rotated Schwefel's function                      & Multimodal    & [-100, 100]     & 10           & 1         \\ \hline
    C19F8        & Shifted and rotated Schaffer's function                      & Multimodal    & [-100, 100]     & 10           & 1         \\ \hline
    C19F9        & Shifted and rotated  Happy Cat function                      & Multimodal    & [-100, 100]     & 10           & 1         \\ \hline
    C19F10       & Shifted and rotated Ackley function                          & Multimodal    & [-100, 100]     & 10           & 1         \\ \hline
    C21F1        & Shifted and Rotated Bent Cigar function                      & Unimodal      & [-100, 100]     & 10           & 100       \\ \hline
    C21F2        & Shifted and Rotated Schwefel's function                      & Basic         & [-100, 100]     & 10           & 1100      \\ \hline
    C21F3        & Shifted and Rotated Lunacek bi-Rastrigin function            & Basic         & [-100, 100]     & 10           & 700       \\ \hline
    C21F4        & Expanded Rosenbrock's plus Griewangk's function              & Basic         & [-100, 100]     & 10           & 1900      \\ \hline
    C21F5        & Hybrid function 1                                            & Hybrid        & [-100, 100]     & 10           & 1700      \\ \hline
    C21F6        & Hybrid function 2                                            & Hybrid        & [-100, 100]     & 10           & 1600      \\ \hline
    C21F7        & Hybrid function 3                                            & Hybrid        & [-100, 100]     & 10           & 2100      \\ \hline
    C21F8        & Composition function 1                                       & Composition   & [-100, 100]     & 10           & 2200      \\ \hline
    C21F9        & Composition function 2                                       & Composition   & [-100, 100]     & 10           & 2400      \\ \hline
    C21F10       & Composition function 3                                       & Composition   & [-100, 100]     & 10           & 2500      \\ \hline
    C22F1        & Shifted and full Rotated Zakharov function                   & Unimodal      & [-100, 100]     & 10           & 300       \\ \hline
    C22F2        & Shifted and full Rotated Rosenbrock's function               & Basic         & [-100, 100]     & 10           & 400       \\ \hline
    C22F3        & Shifted and full Rotated Expanded Schaffer's F7              & Basic         & [-100, 100]     & 10           & 600       \\ \hline
    C22F4        & Shifted and full Rotated Non-Continuous Rastrigin's function & Basic         & [-100, 100]     & 10           & 800       \\ \hline
    C22F5        & Shifted and full Rotated Levy function                       & Basic         & [-100, 100]     & 10           & 900       \\ \hline
    C22F6        & Hybrid function 1                                            & Hybrid        & [-100, 100]     & 10           & 1800      \\ \hline
    C22F7        & Hybrid function 2                                            & Hybrid        & [-100, 100]     & 10           & 2000      \\ \hline
    C22F8        & Hybrid function 3                                            & Hybrid        & [-100, 100]     & 10           & 2200      \\ \hline
    C22F9        & Composition function 1                                       & Composition   & [-100, 100]     & 10           & 2300      \\ \hline
    C22F10       & Composition function 2                                       & Composition   & [-100, 100]     & 10           & 2400      \\ \hline
    C22F11       & Composition function 3                                       & Composition   & [-100, 100]     & 10           & 2600      \\ \hline
    C22F12       & Composition function 4                                       & Composition   & [-100, 100]     & 10           & 2700      \\ \hline
\end{longtable}
\setlength\LTleft     {0pt}
\setlength\LTcapwidth {\linewidth}

\subsection{Real-world problems \label{sec:real_world_problems}}
The following ten \glspl{rwp} have been used to evaluate the applicability of~\gls{ifox} in the real-world:~\gls{bcp},~\gls{cbp},~\gls{csd},~\gls{csp},~\gls{gtp},~\gls{pvd},~\gls{pld},~\gls{srp},~\gls{tcp}, and~\gls{wbp}. 
\begin{itemize}
    \item\gls{bcp}: Minimizes the cost-to-capacity ratio, lightweight structural weight, and maximizes deadweight capacity by optimizing six bulk carrier design variables ($L$, $B$, $D$, $T$, $V_k$, $C_B$), subject to nine nonlinear structural and hydrodynamic constraints.
          
    \item\gls{cbp}: Reduces cantilever beam weight by selecting optimal values for five structural variables, ensuring adequate load-bearing capacity.
          
    \item\gls{csd}: Minimizes spring weight through optimal selection of wire diameter ($d$), coil diameter ($D$), and number of active coils ($N$), satisfying constraints related to shear stress, surge frequency, and deflection.
          
    \item\gls{csp}: Optimizes vehicle structural weight, adhering to stringent side-collision safety standards using eleven mixed-type design variables and nonlinear crashworthiness constraints.
          
    \item\gls{gtp}: Determines integer gear teeth numbers ($T_a$, $T_b$, $T_d$, $T_f$ between 12 and 60) to minimize error from a desired gear ratio of 6.931.
          
    \item\gls{pvd}: Minimizes pressure vessel cost using four decision variables: shell thickness ($T_s$), head thickness ($T_h$), vessel radius ($R$), and cylindrical length ($L$).
          
    \item\gls{pld}: Minimizes oil volume in hydraulic piston mechanisms by optimizing piston components ($H$, $B$, $D$, and $X$) during lever actuation between $0^\circ$ and $45^\circ$.
          
    \item\gls{srp}: Minimizes speed reducer mass through seven design variables: face width ($Z_1$), tooth module ($Z_2$), number of pinion teeth ($Z_3$), shaft lengths ($Z_4$, $Z_5$), and shaft diameters ($Z_6$, $Z_7$), subject to gear and shaft stress constraints.
          
    \item\gls{tcp}: Optimizes tubular column outer diameter ($d$) and thickness ($t$) to minimize material cost, ensuring axial stress and buckling constraints are met.
          
    \item\gls{wbp}: Minimizes welded beam fabrication cost via four variables: weld thickness ($h$), joint length ($l$), beam height ($t$), and beam thickness ($b$), under shear stress, bending stress, buckling, and deflection constraints.
\end{itemize}
Implementations of these \glspl{rwp} are publicly accessible at: \url{https://github.com/thieu1995/enoppy} \cite{pdo_2022,ihaoavoa_2022,moeosma_2023,pvd,pvd_mathmatical}.

\subsection{FOX optimization algorithm\label{sec:fox_methodology}}
\Gls{fox} is a recent optimization algorithm inspired by the hunting behavior of red foxes, it simulates red foxes' behavior in the wild, including walking, jumping, searching for food, and hunting prey~\cite{fox}. It incorporates physics-based principles such as prey detection based on sound and distance, agent jumping governed by gravity and direction, and additional computations such as timing and walking~\cite{foxann}. These features make~\gls{fox} a competitive optimization algorithm, outperforming several established methods such as~\gls{pso},~\gls{ga},~\gls{gwo},~\gls{fdo}, and~\gls{woa}~\cite{fox_tsa}. Furthermore, it has been widely adopted in optimization research due to its effectiveness and robustness~\cite{used_fox_1, used_fox_2, used_fox_3}.

\Gls{fox} operates as a population-based algorithm, where multiple search agents, referred to as~\glspl{fa}, work independently to find the optimal solution. Each agent has its own solution and fitness value, and they collectively strive to achieve the best fitness value in an iterative manner~\cite{fox}. However,~\gls{fa} moves within the problem search space; it always thinks of two options, either exploration or exploitation. The probability of choosing exploration and exploitation in the~\gls{fox} are fixed to 50\% for each process. When the~\gls{fa}  decides to move, the likelihood of exploration is equal to the probability of exploitation, meaning that~\gls{fa} has a probability of 0.5 for exploration and 0.5 for exploitation~\cite{fox_tsa}. The implementation is available at \url{https://github.com/hardi-mohammed/fox}.

\subsubsection{Exploitation in FOX optimization algorithm}
\Glspl{fa} estimate the distance to their prey based on the time it takes for the ultrasound signal to reach them, given that sound travels at a constant speed of 343 meters per second~\cite{soundSpeed}. This estimation process enables the~\glspl{fa} to determine when to jump to capture their prey. The jump mechanism enhances~\gls{fox} 's ability to escape local optima, resulting in superior performance on benchmark test functions and real-world engineering design problems~\cite{fox, optimizationYossra, uot_2}. Consequently, the~\gls{fa} 's success in capturing prey is closely tied to its ability to interpret the sound's travel time while jumping accurately~\cite{used_fox_3}. However,~\gls{fa} exploits through several steps inspired by nature and modeled using mathematical and physical equations.
\begin{equation}
    Sp\_S = \frac{BestX}{Time\_ST} \label{eq:sps}
\end{equation}
\begin{equation}
    Dist\_ST = Sp\_S \odot Time\_ST \label{eq:dist_st}
\end{equation}
where $Dist\_ST$ is the distance of sound travels, $Sp\_S$ is the sound speed in the medium that is approximately $343$ meter per second, but the~\gls{fox} calculates modified $Sp\_S$ based on the Eq~(\ref{eq:sps}), $Time\_ST$ is a random time variable in range $[0,1]$ required by the sound wave to travels from the prey to the~\gls{fa},  and $it$ is the current epoch~\cite{fox}.
\begin{equation}
    Dist\_Fox\_Prey = Dist\_ST \cdot 0.5 \label{eq:dist_fox_prey}
\end{equation}
where $Dist\_Fox\_Prey$ is the distance between the~\gls{fa} and his prey and $BestX$ is the global best solution~\cite{fox}.

Initially, the~\gls{fa} use Eq~(\ref{eq:dist_st}) to calculate the distance of sound that travels from the prey by multiplying the random time variable by the modified speed of the sound from Eq~(\ref{eq:sps}). Red foxes have large ears compared to their head size, which works as radar to enable them to capture the ultrasound waves reflected from the environment. However,~\gls{fa} uses the principle of ultrasound waves inspired by the Doppler effect to determine the prey's location as presented in the Eq~(\ref{eq:dist_fox_prey}).
\begin{equation}
    Jump = 0.5 \cdot 9.81 \cdot t^2, \quad t = \frac{AVG(Time\_ST)}{2} \label{eq:jump}
\end{equation}
where $Jump$ is the control variable used in the following Eq~(\ref{eq:x_exploit}) to determine the new position (solution), $t$ is the average time divided by 2, and the approximate value 9.81 is the gravitational constant, ignoring the effects of air resistance, also termed $G$~\cite{fox}.
\begin{equation}
    X_{it+1} = Dist\_Fox\_Prey \cdot Jump \cdot
    \begin{cases}
        c_1, & \text{if } p>0.18 \\
        c_2, & \text{otherwise}
    \end{cases}
    \label{eq:x_exploit}
\end{equation}

At this moment, the~\gls{fa} has determined the location of the prey using the Eq~(\ref{eq:dist_fox_prey}), and he must decide how he will hunt. Thus,~\gls{fa} has two options: jumping towards the north or the opposite direction using the jump variable from the Eq~(\ref{eq:jump}). The method of determining the direction of the jump depends on a random variable called $p$ in the range $[0, 1]$ and two hyperparameters, $c_1$ and $c_2$, where the value of $c_1$ in range $[0, 0.18]$, and the value of $c_2$ in range $[0.19, 0.82]$, which are fixed numbers tuned at the beginning of the optimization process. Eq~(\ref{eq:x_exploit}) is used to carry out the process of hunting the prey (find the new solution $X_{it+1}$), depend on the variable $p$, which is the probability of using $c_1$ or $c_2$ in hunting process~\cite{fox}.

The exploitation process searches locally to refine the best solution by making slight adjustments multiple times. However, if the optimization algorithm relies on exploitation only, it may become stuck in local optima and cannot discover new solutions. Hence, optimization algorithms should use both exploitation and exploration. On the other hand, the exploration process expands the search globally to find new potential solutions. The following section will discuss the~\gls{fox} 's exploration phase in detail.

\subsubsection{Exploration in FOX optimization algorithm}
During the exploration phase,~\glspl{fa} employs a random walk strategy to locate potential solutions, analogous to how red foxes search for prey. They utilize their ability to detect ultrasound signals to assist in the search process of locating prey~\cite{used_fox_1}. The method of searching is facilitated through controlled random walks, ensuring the~\gls{fa} progresses toward the prey while maintaining an element of randomness. During this phase,~\gls{fox} 's unique random walk and distance measurement mechanisms enable a refined search process, distinguishing it from other established swarm-based algorithms like~\gls{pso} and~\gls{gwo}~\cite{used_fox_3}.
\begin{equation}
    a = 2 \cdot (it - (\frac{1}{Max}))  \label{eq:a}
\end{equation}
\begin{equation}
    Mint = Min(tt), \quad tt = \frac{\sum_{}^{}Time\_ST}{dim} \label{eq:mint}
\end{equation}
\begin{equation}
    X_{it+1} = BestX \odot rnd(1, dim) \cdot Mint \cdot a \label{eq:x_explore}
\end{equation}
where $rnd(1, dim)$ is a uniform random value used to provide slight perturbation for the solution to enhance the diversity, and $dim$ is the problem~\cite{fox}.

As mentioned, the~\gls{fa} explore the problem search space by searching for the best solution by random walk strategy. The Eq~(\ref{eq:x_explore}) used for exploration to search for new position ($X_{it+1}$) for the current~\gls{fa}. Additionally, it has two adaptive hyperparameters: $a$ is calculated based on the~\gls{fox} epochs as seen in Eq~(\ref{eq:a}) and $Mint$ is calculated based on the minimum averaged time as seen in the Eq~(\ref{eq:mint}).

\begin{figure}[htb]
    \centering
        \includegraphics[width=0.5\textwidth]{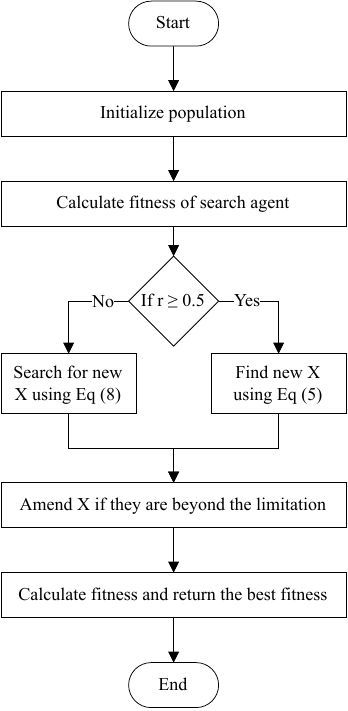}

    \caption{Flowchart of FOX optimization algorithm~\cite{fox}}
    \label{fig:fox_algorithm}
\end{figure}

After each epoch of optimization (whether exploration or exploitation), the~\gls{fa} evaluates the objective function value (fitness) based on the current solution. If this solution leads to better convergence, it is considered the best solution. \Cref{fig:fox_algorithm} visualizes the primary steps of~\gls{fox} redrawn from the original study~\cite{fox}.

\subsubsection{limitations in FOX optimization algorithm\label{sec:fox_limitations}}
\begin{figure}[htb]
    \centering
        \includegraphics[width=0.7\textwidth]{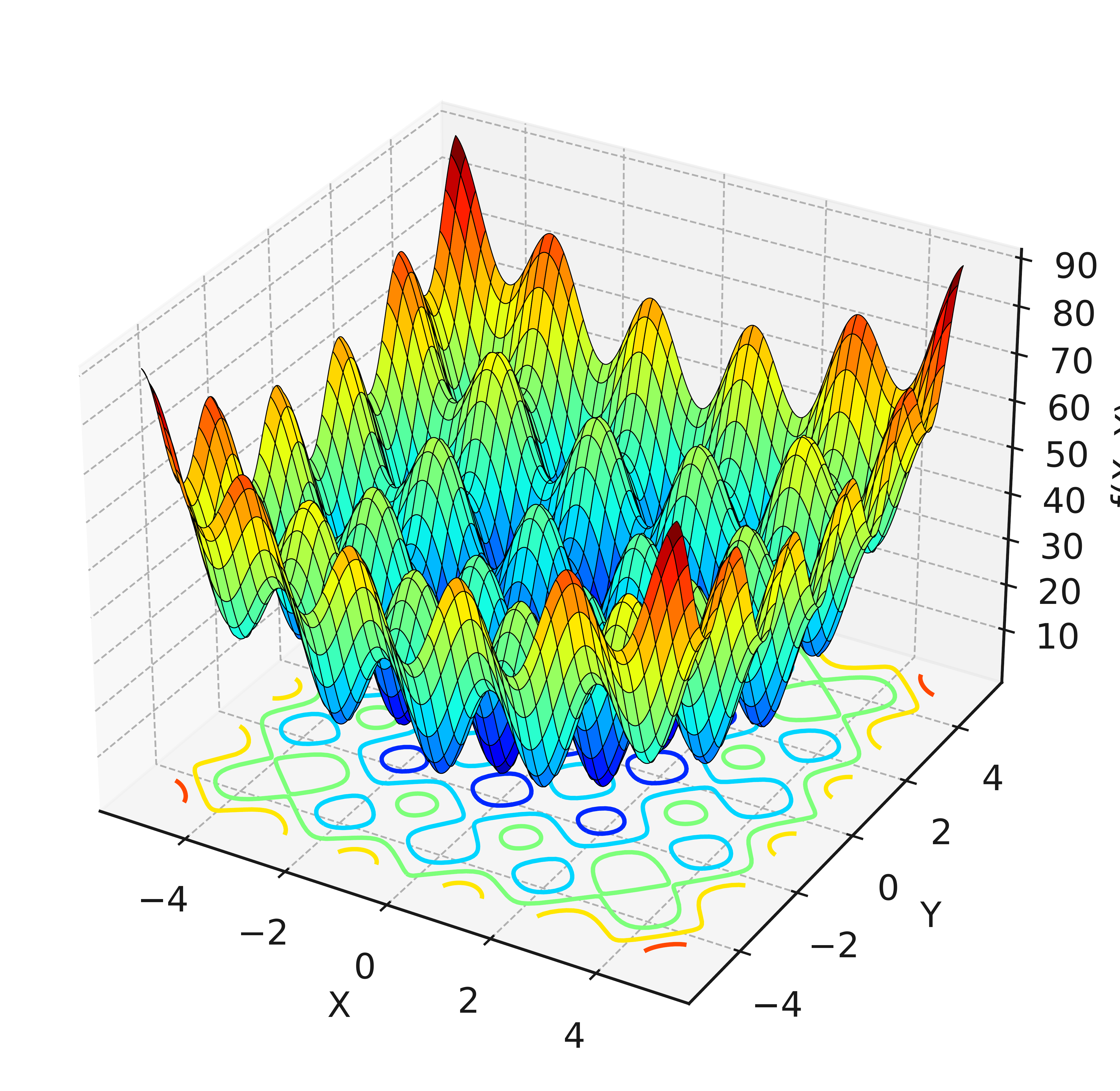}
    \caption{ 3D visualization of the Rastrigin function from the~\gls{cec} 2019 benchmark test functions. It contains a large number of local optima. The optimization in such a function (landscape) requires a balance between exploration to avoid local optima and exploitation to converge accurately near the global optimum. It is important to use adaptive methods to balance exploration and exploitation in such situations. Adopted from~\cite{rastrigin_figure}.}
    \label{fig:rastrigin_function}
\end{figure}

Although the~\gls{fox} is superior to several established optimization algorithms, it has notable limitations, as observed through experiments conducted in this paper across various tests. Furthermore, the authors of the~\gls{fox} acknowledged these limitations, stating: "\textit{The FOX exploration phase can be enhanced to be more competitive against other algorithms. The exploitation phase of the FOX also can be improved by changing the equations used to find the distance of the red Fox from prey and the jump equation. The convergence of the fitness value can be improved based on changing exploration and exploitation equations}" ~\cite{fox}. Additionally, the method of exploitation and exploration in the~\gls{fox} is static, with each being set to 0.5. This does not match the nature of many optimization problems, as some issues need more exploration, and others need more exploitation. For instance, the search space has many local optima in a complex benchmark test function C19F4, as visualized in Fig~\ref{fig:rastrigin_function}. Therefore, more exploration is required in the early stages. However, with the fixed 0.5 value, the algorithm may start exploiting too early and become trapped in local optima. Conversely, in simple problems, excessive exploration may result in inefficient convergence. This shows that using a fixed value limits the algorithm's ability to adapt to the nature of the problem.

\subsection{Improved FOX optimization algorithm\label{sec:impooved_fox_methodology}}
Several significant improvements have been made to the~\gls{fox}. Address limitations, contribute to higher performance, and reduce the restriction the~\gls{fox} faces. Consequently, a new adaptive method for balancing exploration and exploitation will be discussed in~\Cref{sec:adaptive_exploitation_exploration} section, modifying certain equations and rewriting others for better clarity. Furthermore, the complete implementation is available at~\url{https://github.com/mwdx93/ifox}.

\subsubsection{Exploitation in the improved FOX optimization algorithm\label{sec:ifox_exploitation_ch3}}
Some original equations in~\gls{fox} are modified throughout the exploitation phase due to their limitations and ambiguity, particularly in time and distance computations. The reasons for these modification such as reducing the number of equations and achieving higher performance with the~\gls{ifox}, are discussed below. Additionally, the~\gls{fox} 's equations are slightly improved, but not entirely, to preserve their general structure and main idea, maintaining their superiority.
\begin{equation}
    Dist\_Fox\_Prey = \frac{BestX}{\cancel{Time\_ST}} \cdot \cancel{Time\_ST} \cdot 0.5 \label{eq:dist_fox_prey_full}
\end{equation}
\begin{equation}
    Dist\_Fox\_Prey = BestX \cdot 0.5\label{eq:ifox_dist}
\end{equation}

Eq~(\ref{eq:dist_fox_prey_full}) represents the direct substitution of Eq~(\ref{eq:sps}) and (\ref{eq:dist_st}) into Eq~(\ref{eq:dist_fox_prey}). When expanded, it becomes evident that the time variable $Time\_ST$ cancels out, leading to a redundant form where the distance to prey becomes directly proportional to the best solution $BestX$, scaled by a constant factor. This cancellation introduces ambiguity in the~\gls{fox} exploitation mechanism, as the $Time\_ST$ variable initially intended to simulate prey detection based on sound travel time becomes mathematically ineffective.

To resolve this ambiguity, Eq~(\ref{eq:ifox_dist}) is introduced in the~\gls{ifox}. It simplifies the computation by directly defining the distance between the fox agent and the prey as half of the best solution $BestX$. This modification eliminates the unnecessary dependence on the canceled $Time\_ST$ variable, resulting in a clearer, more consistent mathematical model. Additionally, it maintains the physical intuition behind the exploitation phase while enhancing computational efficiency and clarity. Thus, the transition from Eq~(\ref{eq:dist_fox_prey_full}) to Eq~(\ref{eq:ifox_dist}) represents a meaningful refinement in the~\gls{ifox} design. Additionally, in the original~\gls{fox}, the jump in Eq~(\ref{eq:jump}) is computed dynamically during each epoch based on the time variable and the gravitational constant. In contrast, in the~\gls{ifox}, half of the gravitational constant value is precomputed and defined as a constant outside the main optimization loop. This refinement enhances computational clarity, reduces redundancy, and improves the overall readability of the optimization process.

The~\gls{ifox} utilized a new adaptive value, \textit{alpha} $\alpha$. It plays a crucial role in most of the~\gls{ifox} 's steps because of its impact on the improved exploration and exploitation equations. It must be updated adaptively at each epoch using Eq~(\ref{eq:alpha_calculation}). However, the alpha $\alpha$ value has been employed to enhance the diversity and potentially enhance solutions for improved performance. Similarly, the alpha $\alpha$ value is also responsible for computing a random uniform value, denoted as \textit{beta} $\beta$, which will be used in the modified exploration and exploitation equations to improve them further. The following equations describe how to compute the alpha $\alpha$ value in the primary loop and beta $\beta$ vector in the~\gls{fa} loop, respectively:
\begin{equation}
    \alpha = \alpha_{min} + (1 - \alpha_{min}) \cdot (1 - \frac{it}{epochs}) \label{eq:alpha_calculation}
\end{equation}
\begin{equation*}
    \alpha_{min} = \frac{1}{0.5 \cdot epochs}
\end{equation*}
\begin{equation}
    \beta =
    \begin{cases}
        \text{LevyFlight}(dim) \cdot \alpha, & \text{if } \text{rnd}(0,1) < \alpha \\
        \text{rnd}(-\alpha, \alpha, dim),    & \text{otherwise}
    \end{cases}
    \label{eq:beta_calculation_ifox}
\end{equation}
where $\alpha_{min}$ denotes the empirically chosen lower bound on, ensuring a minimal nonzero perturbation throughout the optimization process. In the~\gls{ifox}, the~$\beta$ vector is adaptively generated based on $\alpha$. Specifically, if a random number in $[0,1]$ is less than $\alpha$, the~$\beta$ is computed using a Lévy flight distribution, random walk with mostly small steps and occasional large jumps~\cite{levy}, scaled by $\alpha$; otherwise, it is drawn uniformly from the range $[-\alpha, \alpha]$ for each dimension. 

Eq~(\ref{eq:beta_calculation_ifox}) is employed to compute a crucial vector in~\gls{ifox}, which is generated adaptively at each epoch within the~\gls{fa} loop and subsequently used in the improved exploitation equation: 
\begin{equation}
    Xt = Dist\_Fox\_Prey \odot \beta \cdot Jump \label{eq:x_exploit_ifox}
\end{equation}
In the original~\gls{fox}, the exploitation step is computed based on two fixed quantities: the distance between the fox and its prey from the Eq~(\ref{eq:dist_fox_prey}) and the jump value from Eq~(\ref{eq:jump}). These values remain constant for each~\gls{fa} during an epoch and are updated only once per main loop epoch. Additionally, the constants $c_1$ and $c_2$ used in the movement decision process are static and do not vary. This static behavior limits the diversity of solutions and increases the risk of premature convergence. In contrast, Eq~(\ref{eq:x_exploit_ifox}) introduces adaptivity by incorporating the$\beta$ vector, which is recalculated for each agent in every epoch. The dynamic nature of~$\beta$ significantly enhances solution diversity during the exploitation, leading to better search capabilities and faster convergence.

\subsubsection{Exploration in the improved FOX optimization algorithm\label{sec:ifox_exploration_ch3}}
The analysis of the~\gls{fox}'s exploration solution produced by Eq~(\ref{eq:x_explore}) revealed the need for improvements. The exploration process should generate solutions with acceptable diversity; however, the~\gls{fox} exploration produces solutions with massive variations. For instance, when testing the~\gls{fox} using a classical function, the best solution is $BestX = [0.41, -2.3]$, while $BestX_{it+1} = [12.7, -9.5]$, which is significantly different. Hence, the~\gls{ifox} replaces the multiplication process with an addition process, ensuring a balance by generating solutions that exhibit sufficient—but not excessive—diversity. Furthermore, the variables $a$ and $Mint$ have been replaced with the proposed alpha $\alpha$ and beta $\beta$, respectively. These changes lead to the production of exploration solutions $Xr$ that outperform those generated the~\gls{fox}, as presented in the improved equation below:
\begin{equation}
    Xr = BestX + \beta \cdot \alpha \label{eq:x_explore_ifox}
\end{equation}

\begin{figure}[htb]
    \centering
        \includegraphics[width=0.9\textwidth]{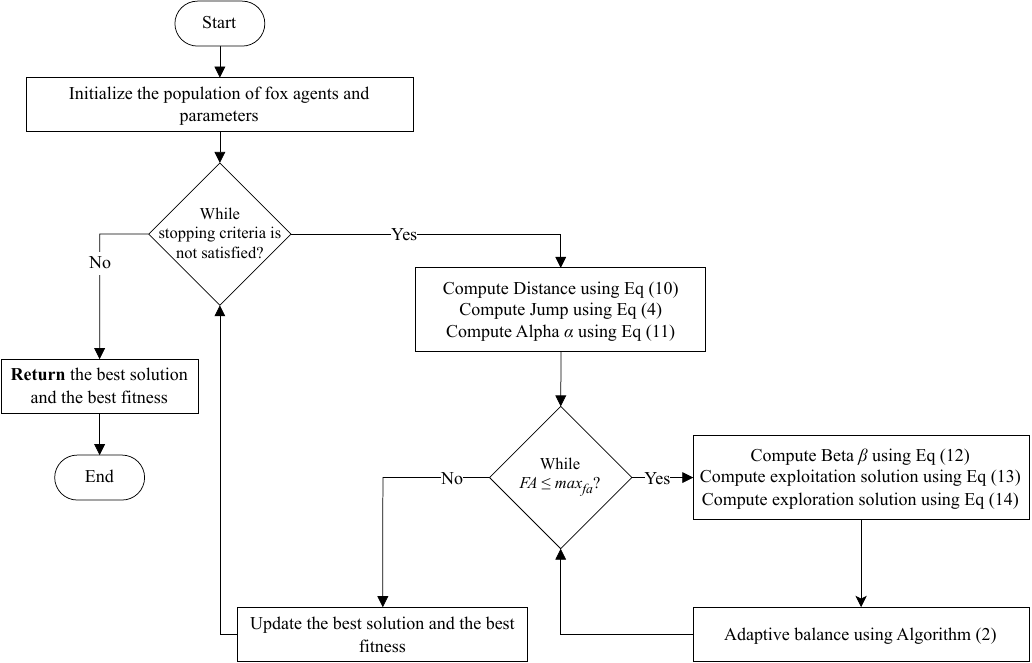}
    \caption{Flowchart of improved FOX optimization algorithm }
    \label{fig:ifox_flowchart}
\end{figure}

\begin{algorithm}[htb]
    \caption{Main loop of the Improved FOX optimization algorithm\label{algo:ifox_main}}
    \begin{algorithmic} 
        \State Initialize $max\_iter, max_{fa}, Jump, BestX, BestFitness$
        \State Initialize the fox agents population $ X = \{ X_{fa} \mid fa = 1, 2, \dots, max_{fa} \} $
        \State \textbf{constant} $half\_G \leftarrow 4.905$
        \While{$it < max\_iter$}
        \State Compute the distance \Comment Eq~(\ref{eq:ifox_dist})
        \State $Jump \leftarrow half\_G \times t^2$ \Comment Rewritten Eq~(\ref{eq:jump})
        \State Compute alpha $\alpha$ \Comment Eq~(\ref{eq:alpha_calculation})
        \For{$fa = 0$ to $max_{fa}$}
        \State Compute beta $\beta$ \Comment Eq~(\ref{eq:beta_calculation_ifox})
        \State Compute exploitation solution $X_t$ \Comment Eq~(\ref{eq:x_exploit_ifox})
        \State Compute exploration solution $X_r$ \Comment Eq~(\ref{eq:x_explore_ifox})
        \State $X_{fa} \leftarrow$ \Call{Balance}{$BestX$, $X_t$, $X_r$, $\beta$} \Comment Algorithm~\ref{algo:ifox_balance}
        \EndFor
        \For{$fa = 0$ to $max_{fa}$}
        \State $f \leftarrow Fitness(X_{fa})$
        \If{$f \leq BestFitness$}
        \State $BestFitness \leftarrow f$
        \State $BestX \leftarrow X_{fa}$
        \EndIf
        \EndFor
        \State $it \leftarrow it + 1$
        \EndWhile
        \State \Return $BestX$, $BestFitness$
    \end{algorithmic}
\end{algorithm}

\subsubsection{Adaptive exploration and exploitation \label{sec:adaptive_exploitation_exploration}}
\begin{figure}[htb]
    \centering 
        \includegraphics[width=0.5\textwidth]{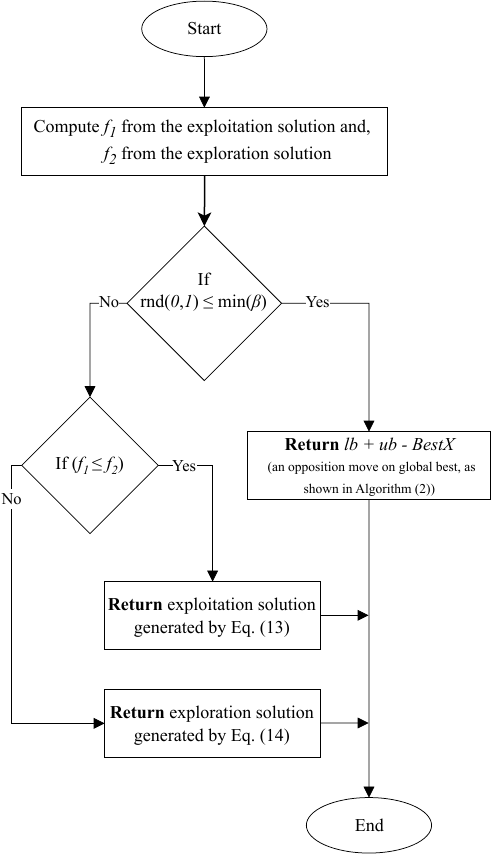}
    \caption{Adaptive exploration and exploitation method of Improved FOX optimization algorithm}
    \label{fig:ifox_balance}
\end{figure}

This section explains how the method of balancing exploration and exploitation is transformed into an adaptive approach in the~\gls{ifox}, based on the fitness values of the problem solutions inspired by the study in~\cite{fdo}. The~\gls{fox} uses a random variable, $r$, which ranges between $0$ and $1$, to determine whether the algorithm will perform exploration or exploitation. Specifically, if $r \geq 0.5$, the~\gls{fa} performs exploitation; otherwise, it performs exploration. In contrast, the~\gls{ifox} eliminates the random $r$ variable and instead uses the fitness values of candidate solutions, as visualized in Fig~\ref{fig:ifox_balance}. The proposed method adaptively selects exploration or exploitation based on the solution's fitness value while incorporating a small probability of performing an opposition-based move.

In Algorithm~\ref{algo:ifox_balance}, after computing the fitness values $f_1$ and $f_2$, a random number in $[0,1]$ is generated and compared to $\min(\beta)$. Suppose the random number is less than or equal to the minimum of beta $\beta$. In that case, the FA performs an opposition move by replacing the current solution with an opposite position relative to the global best. This adaptive strategy introduces controlled randomness into the search process: when $\beta$ is large, more opposition moves occur to promote exploration; when $\beta$ is small, the search process favors exploitation. This mechanism further improves the~\gls{ifox}'s ability to escape local optima and enhances overall convergence behavior. However, if the opposition move is not triggered, the~\gls{ifox} compares the two fitness values, $f_1$ and $f_2$, obtained by evaluating the exploitation solution from Eq~(\ref{eq:x_exploit_ifox}) and exploration solution from Eq~(\ref{eq:x_explore_ifox}), respectively. If $f_1$ is less than $f_2$ (for minimization problems), the solution from the exploitation phase is selected for the current~\gls{fa}; otherwise, the exploration solution is chosen. This adaptive decision-making process enhances the algorithm's ability to balance exploration and exploitation more effectively than the static balancing method used in the~\gls{fox}, leading to faster convergence and higher-quality solutions.

\begin{algorithm}[htb]
    \caption{Adaptive method for exploration and exploitation\label{algo:ifox_balance}}
    \begin{algorithmic} 
        \Function{Balance}{$BestX$, $X_t$, $X_r$, $\beta$}
        \State $f_1 \leftarrow Fitness(X_t)$
        \State $f_2 \leftarrow Fitness(X_r)$
        \If{$\text{rnd}(0,1) \leq \text{min}(\beta)$}
        \State \Return $lb + ub - BestX$ \Comment $lb$ and $ub$ are the problem's bounds.
        \Else
        \If{$f_1 \leq f_2$}
        \State \Return $X_t$
        \Else
        \State \Return $X_r$
        \EndIf
        \EndIf
        \EndFunction
    \end{algorithmic}
\end{algorithm}

\section{Results\label{sec:results}}
Experimental evaluations were carried out to assess the performance of the proposed~\gls{ifox}. Its results were compared against several recently improved and~\gls{sota} optimization algorithms discussed in the~\Cref{sec:literature_review}, including~\gls{alshade},~\gls{bboa},~\gls{bda},~\gls{leo},~\gls{cma-es},~\gls{coa},~\gls{cpo},~\gls{fox},~\gls{ho},~\gls{iaga},~\gls{shoa},~\gls{ipso},~\gls{iwoa},~\gls{lshade},~\gls{nro}, and~\gls{vw-gwo}. They were implemented in Python and executed on an MSI GL63 8RD laptop, equipped with an Intel\textsuperscript{\textregistered} Core\texttrademark{} i7-8750H CPU (12 threads) and 32GB of RAM. The experimental settings---including the number of epochs, population size, and trials---were standardized to ensure consistency, as detailed in Table\ref{tab:parameters_settings}. The evaluation employed a comprehensive set of 81 benchmark test functions and ten \glspl{rwp}, with each experiment independently executed 30 times under identical conditions. Results were aggregated by averaging to ensure the robustness and reliability of the performance assessment across diverse optimization tests.
\begin{table}[htb]
    \caption{General parameters setting.\label{tab:parameters_settings}}
    \begin{tabularx}{\textwidth}{|X|X|}
        \hline
        \textbf{Parameter}      & \textbf{Value} \\
        \hline
        No. of epochs           & 500            \\ \hline
        No. of populations      & 30             \\ \hline
        No. of trials           & 30             \\ \hline
        Trials aggregation type & Average        \\ \hline
    \end{tabularx}
\end{table}

The performance metrics, including the average convergence (Avg), standard deviation (Std), best convergence (Best), worst convergence (Worst), Bayesian rank-sum test (BRT), and Rank have been used for evaluations. Avg represents the mean best objective value (fitness) over 30 independent runs, indicating the optimization efficiency. Std measures the stability across runs, with lower values implying higher robustness. Best and Worst denote the minimum and maximum objective values achieved, respectively. The BRT provides posterior probabilities of one algorithm outperforming another, addressing issues in multiple comparisons. It counts wins/ties/losses (+/=/-) for each optimization algorithm across the optimization experiment. A win count means better performance than other algorithms, a tie count means equivalent performance, and a loss count means worse performance. The Rank is computed using the Friedman statistical test. It defines the ranking of each optimization algorithm based on the Avg metric across all tested tasks. The lower rank values indicate better overall performance. Furthermore, the~\gls{pt} is the average execution time the optimization algorithm takes per epoch. 

Table~\ref{tab:results_ifox} provides the comprehensive performance results among 17 recent, improved,~\gls{sota} optimization algorithms, including the~\gls{ifox}. The results demonstrate that~\gls{ifox} achieved outstanding performance across various benchmark test functions, consistently ensuring top rankings on several functions like: CL1, CL5, CL8, CL10, CL11, CL13, CL15, CL17, C17F1, C17F2, C17F16, C17F20, C19F1, C19F2, C19F6, C21F9, C22F1, and C22F11. Moreover, in \glspl{rwp},~\gls{ifox} showed strong performance on the CBP and GTP. Total results indicate that~\gls{ifox} achieved 880 wins, 228 ties, and 348 losses across all experiments. It also maintained a promising mean rank of 5.92 and demonstrated computational performance with an average~\gls{pt} of 0.0038 seconds. Finally, these findings emphasize the competitiveness and effectiveness of the proposed~\gls{ifox}, making it comparable or superior to~\gls{sota} and recent optimization algorithms such as~\gls{alshade},~\gls{lshade},~\gls{cma-es}, and~\gls{nro}.

\newgeometry{top=0in}
\renewcommand{\footrule}{\hrule height 0pt \vspace{0mm}}
\begin{landscape}
    \setlength\LTcapwidth {\dimexpr\textwidth+3.5in\relax}
    \renewcommand{\arraystretch}{1.45} 
    \setlength{\tabcolsep}{1pt} 
    \tiny
\begin{longtable}{|l|l|l|l|l|l|l|l|l|l|l|l|l|l|l|l|l|l|l|}
        \caption{The performance results on benchmark test functions and real-world problems \label{tab:results_ifox}}                                                                                                                                                                                                                                                                                                                               \\
        \hline
        \textbf{TID}                                   & \textbf{Metric}                                       & \multicolumn{17}{c|}{\textbf{Optimization Algorithm}}                                                                                                                                                                                                                                                                               \\
        \cline{3-19}
                                                       &                                                       & \textbf{IFOX}                                         & \textbf{ALSHADE} & \textbf{BBOA} & \textbf{BDA}  & \textbf{LEO}  & \textbf{CMAES} & \textbf{COA} & \textbf{CPO} & \textbf{FOX} & \textbf{HO}   & \textbf{IAGA}    & \textbf{SHOA} & \textbf{IPSO} & \textbf{IWOA}   & \textbf{LSHADE} & \textbf{NRO}   & \textbf{VWGWO} \\
        \hline
        \endfirsthead
        
        \multicolumn{19}{@{}l}{\textit{\ldots continued from previous page}}                                                                                                                                                                                                                                                                                                                                                                         \\
        \hline
        \textbf{TID}                                   & \textbf{Metric}                                       & \multicolumn{17}{c|}{\textbf{Optimization Algorithm}}                                                                                                                                                                                                                                                                               \\
        \cline{3-19}
                                                       &                                                       & \textbf{IFOX}                                         & \textbf{ALSHADE} & \textbf{BBOA} & \textbf{BDA}  & \textbf{LEO}  & \textbf{CMAES} & \textbf{COA} & \textbf{CPO} & \textbf{FOX} & \textbf{HO}   & \textbf{IAGA}    & \textbf{SHOA} & \textbf{IPSO} & \textbf{IWOA}   & \textbf{LSHADE} & \textbf{NRO}   & \textbf{VWGWO} \\
        \hline
        \endhead
        
        \multicolumn{19}{r@{}}{\textit{continued on next page \ldots}}                                                                                                                                                                                                                                                                                                                                                                               \\
        \endfoot    
        \endlastfoot
        
        \multicolumn{ 1}{|c|}{\textbf{CL1}}            & Avg                                                   & -2.00E+02                                             & -2.00E+02        & -1.99E+02     & -2.00E+02     & -1.89E+02      & -2.00E+02      & -2.00E+02    & -2.00E+02    & -2.00E+02    & -2.00E+02     & -1.99E+02        & -2.00E+02        & -2.00E+02     & -2.00E+02       & -2.00E+02       & -2.00E+02      & -2.00E+02      \\ \cline{ 2- 19}
        \multicolumn{ 1}{|l|}{}                        & Std                                                   & 0.00E+00                                              & 5.78E-14         & 1.05E+00      & 0.00E+00      & 5.63E+00       & 0.00E+00       & 0.00E+00     & 0.00E+00     & 0.00E+00     & 0.00E+00      & 5.01E-01         & 0.00E+00         & 9.35E-14      & 1.49E-14        & 0.00E+00        & 0.00E+00       & 0.00E+00       \\ \cline{ 2- 19}
        \multicolumn{ 1}{|l|}{}                        & Best                                                  & -2.00E+02                                             & -2.00E+02        & -2.00E+02     & -2.00E+02     & -1.99E+02      & -2.00E+02      & -2.00E+02    & -2.00E+02    & -2.00E+02    & -2.00E+02     & -2.00E+02        & -2.00E+02        & -2.00E+02     & -2.00E+02       & -2.00E+02       & -2.00E+02      & -2.00E+02      \\ \cline{ 2- 19}
        \multicolumn{ 1}{|l|}{}                        & Worst                                                 & -2.00E+02                                             & -2.00E+02        & -1.95E+02     & -2.00E+02     & -1.76E+02      & -2.00E+02      & -2.00E+02    & -2.00E+02    & -2.00E+02    & -2.00E+02     & -1.98E+02        & -2.00E+02        & -2.00E+02     & -2.00E+02       & -2.00E+02       & -2.00E+02      & -2.00E+02      \\
\hline

    \end{longtable}
\end{landscape}
\renewcommand{\footrule}{\hrule height 2pt \vspace{2mm}}
\restoregeometry

\subsection{Convergence results}
Convergence in optimization refers to an algorithm's capability to progressively approach the optimal or near-optimal solution over epochs. The following figures illustrate the convergence performance of~\gls{ifox} (red) compared to other optimization algorithms that are considered in this paper. Each figure displays selected functions where significant differences in convergence are observed.

\begin{figure}[htp]
        \centering
        \includegraphics[width=1\textwidth]{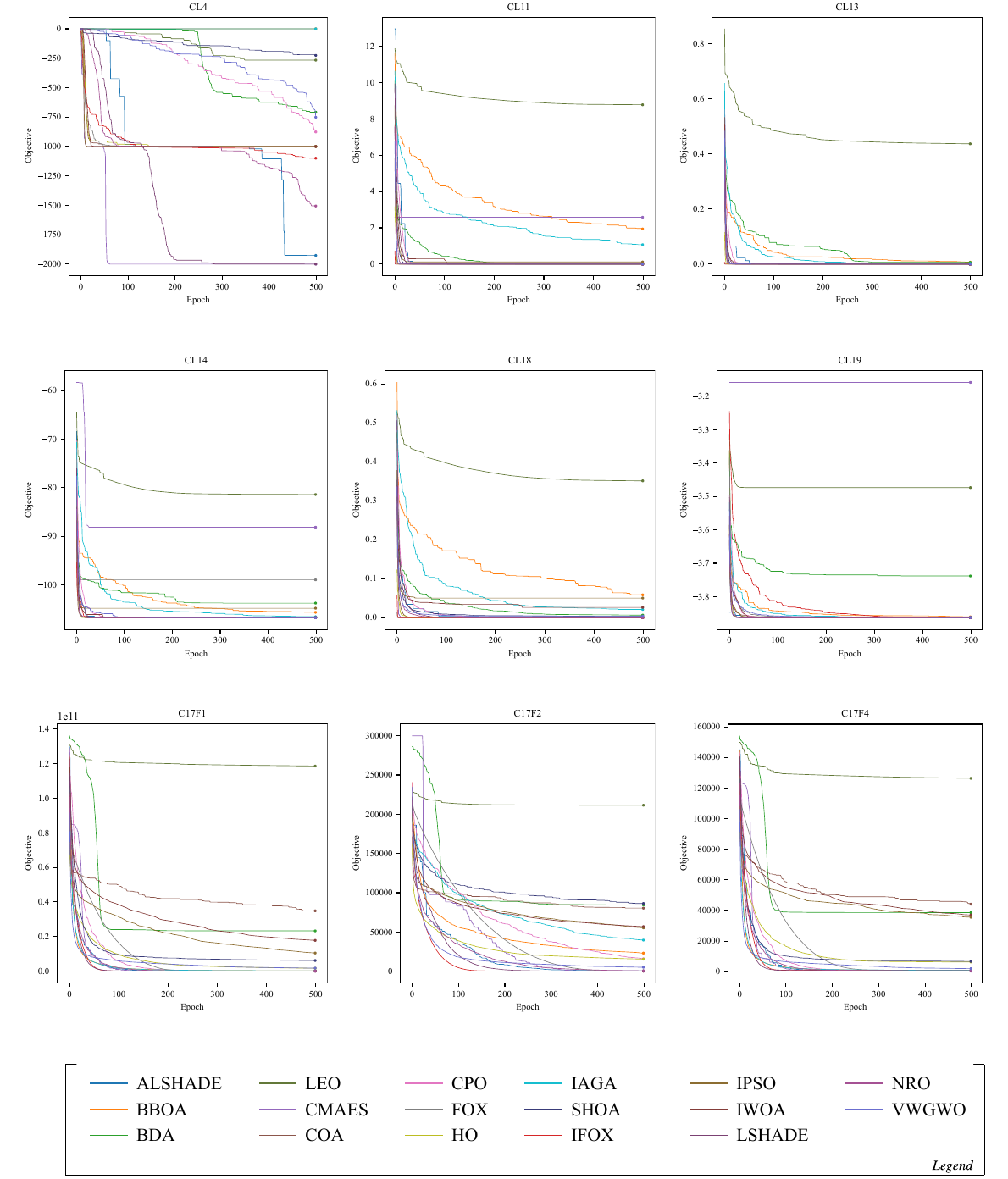}
        \caption{Convergence curves on the functions CL4, CL11, CL13, CL14, CL18, CL19, C17F1, C17F2, and C17F4.}
        \label{fig:convergence_g1}
\end{figure}

\begin{figure}[htp]
        \centering
        \includegraphics[width=1\textwidth]{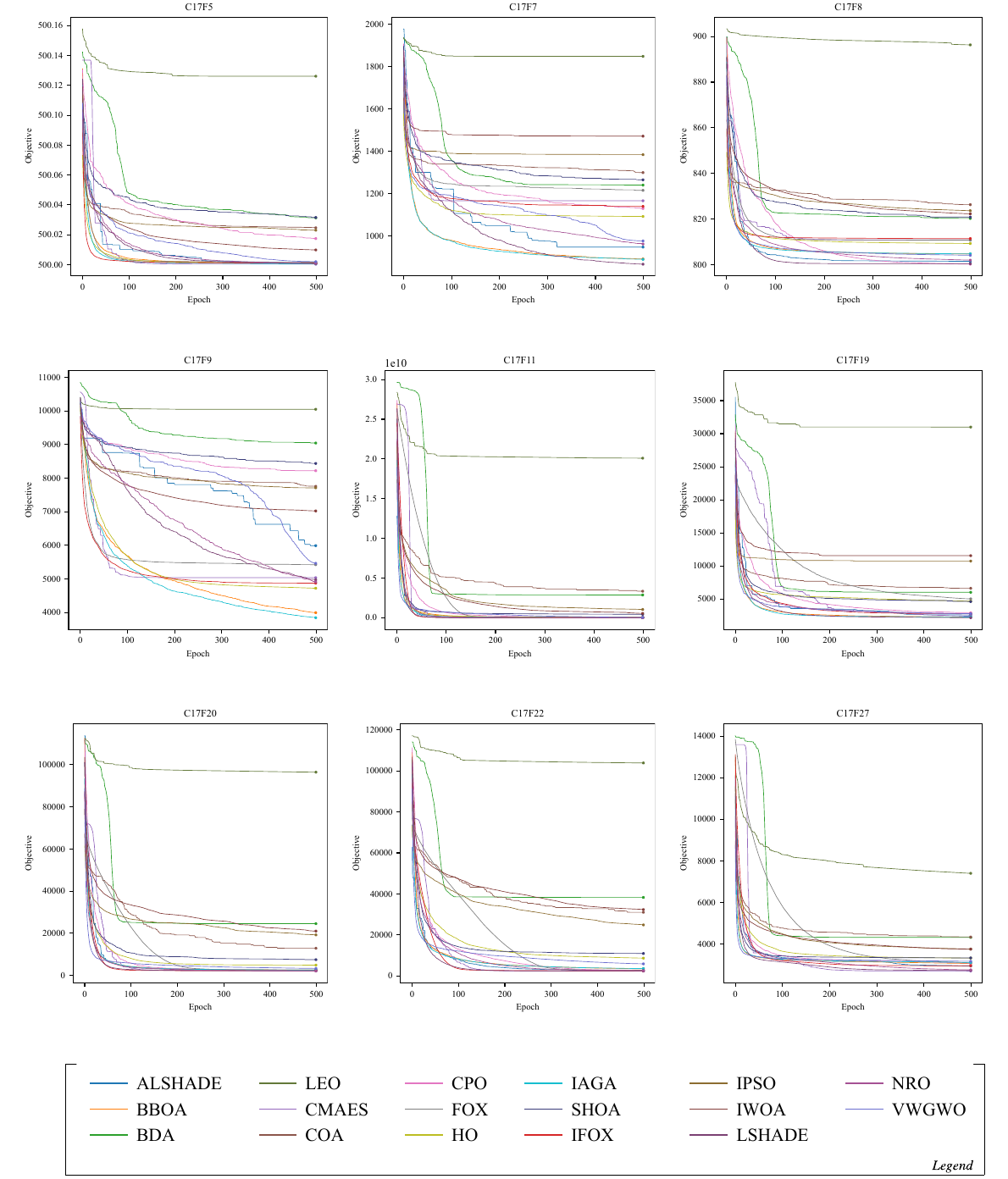}
        \caption{Convergence curves on the functions C17F5, C17F7, C17F8, C17F9, C17F11, C17F19, C17F20, C17F22, and C17F27.}
        \label{fig:convergence_g2}
\end{figure}

\begin{figure}[htp]
        \centering
        \includegraphics[width=1\textwidth]{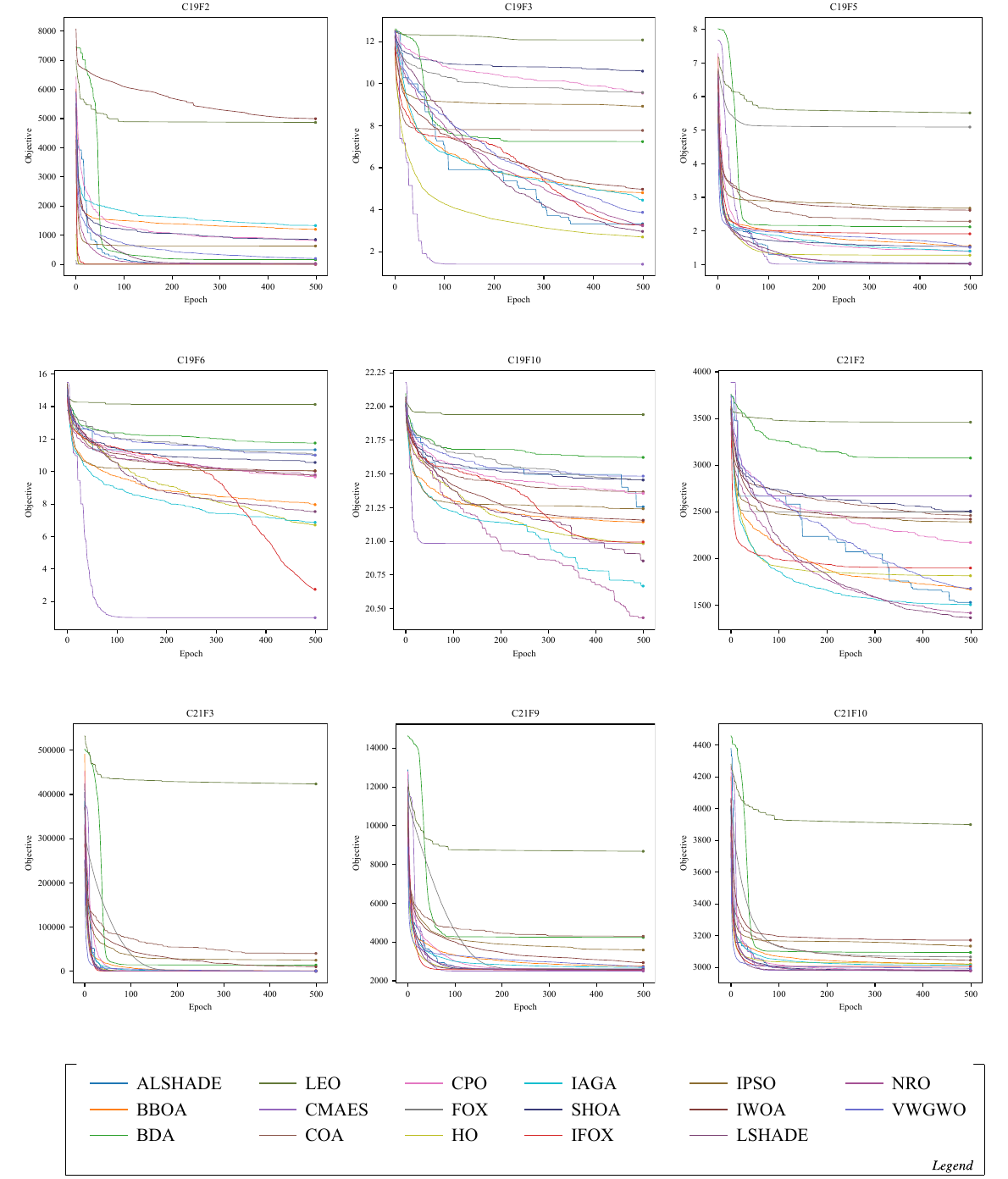}
        \caption{Convergence curves on the functions C19F2, C19F3, C19F5, C19F6, C19F10, C21F2, C21F3, C21F9, and C21F10.}
        \label{fig:convergence_g3}
\end{figure}

\begin{figure}[htp]
        \centering
        \includegraphics[width=1\textwidth]{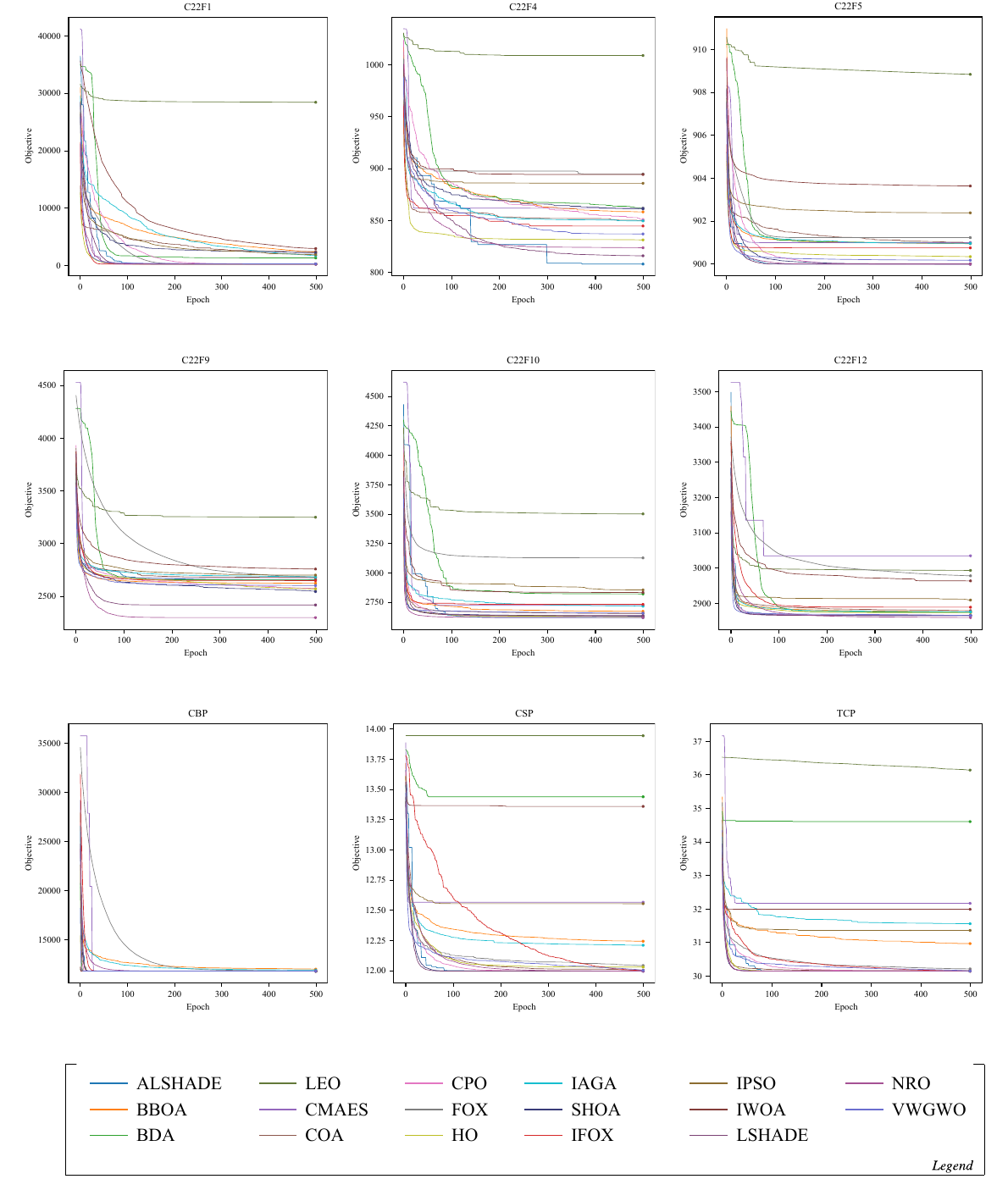}
        \caption{Convergence curves on the functions C22F1, C22F4, C22F5, C22F9, C22F10, C22F12, CBP, CSP, and TCP.}
        \label{fig:convergence_g4}
\end{figure}

The convergence performance of~\gls{ifox} in Figures~\ref{fig:convergence_g1}, \ref{fig:convergence_g2} \ref{fig:convergence_g3} and \ref{fig:convergence_g4}, showed rapid convergence on functions CL11, CL13, CL14, and CL18, outperforming the competing optimization algorithms during the initial epochs (the first ten). Furthermore, it consistently outperformed other algorithms within the first 100 epochs on C17F1, C17F2, C17F4, C17F5, C17F20, C21F9, C21F3, C19F2, and C22F1, reflecting unique convergence. In contrast, for functions C17F7, C17F8, C17F9, C17F11, C17F19, C21F10, C22F9, and C22F10, it demonstrated convergence performance comparable to other algorithms, showing neither a distinct advantage nor significant delayed.
For another group of functions, such as CL4, CL19, C19F3, C19F6, C19F5, C19F10, C21F2, C22F4, and C22F5, the convergence performance of it was relatively weaker, without clear superiority, despite achieving high rankings in terms of final optimal values, as observed in C19F6. Additionally, it demonstrated convergence patterns similar to competing algorithms in real-world optimization problems, except for the~\gls{cbp},~\gls{csp}, and~\gls{tsp} problems, where its convergence was less effective, particularly noticeable in the~\gls{csp} problem.
The above convergence examination shows that~\gls{ifox} exhibits robust convergence performance on several experiments, particularly during early epochs. However, it performed comparably or somewhat weaker on other functions and \glspl{rwp}, making it a good avenue for future improvement. Moreover, further interpretation and analysis of results are required to identify the strengths and limitations of the~\gls{ifox}. Hence, an in-depth analysis will be discussed in the next section.

\section{Discussion\label{sec:discussion}}
This section analyzes and discusses the performance of~\gls{ifox} in comparison with other optimization algorithms. While the results indicate that~\gls{ifox} is generally effective in solving optimization test functions and problems, achieving optimal solutions alone is not sufficient, as many algorithms are capable of this. Therefore, it is essential to examine and compare the convergence behavior, particularly the average convergence metrics, to highlight the advantages of~\gls{ifox} and justify the proposed contributions. As shown in Table~\ref{tab:results_ifox},~\gls{ifox} outperformed most competing algorithms across 15 benchmark test functions and two~\glspl{rwp}. To ensure a rigorous comparison, statistical significance was assessed using the non-parametric Friedman and Wilcoxon signed-rank tests~\cite{friedman_test,statistical_analysis}. A p-value below 0.05 was considered statistically significant in this analysis.

The~\gls{ifox} convergence curves (red) in Figures~\ref{fig:convergence_g1}–\ref{fig:convergence_g4} demonstrate that the algorithm rapidly achieves optimal solutions on CL11, CL13, C17F1, and C17F2 functions before its competitors during initial epochs. The fast exploration and exploitation capabilities arise from the proposed parameters alpha $\alpha$ and beta $\beta$ in Eq~(\ref{eq:alpha_calculation}) and (\ref{eq:beta_calculation_ifox}) that utilized in the proposed adaptive balance strategy. \Gls{ifox} converged similarly to other algorithms on functions such as C17F7 and C22F9 because the problem domains did not give it an advantage during exploration. The convergence time was slower on C19F6 and C22F4 because these functions exhibited short periods of stability and restricted early solution exploration, but~\gls{ifox} proved able to discover efficient final solutions. The performance consistency in \glspl{rwp} showed exceptions only in~\gls{cbp},~\gls{csp}, and~\gls{tcp} because these problems had complex constraints that reduced the effectiveness of~\gls{ifox}. This evidence reveals that~\gls{ifox} functions optimally in smooth continuous domains; however, it still lacks enhancement for highly constrained and discrete problem types.

\begin{figure}[H]
        \centering
        \includegraphics[width=1\textwidth]{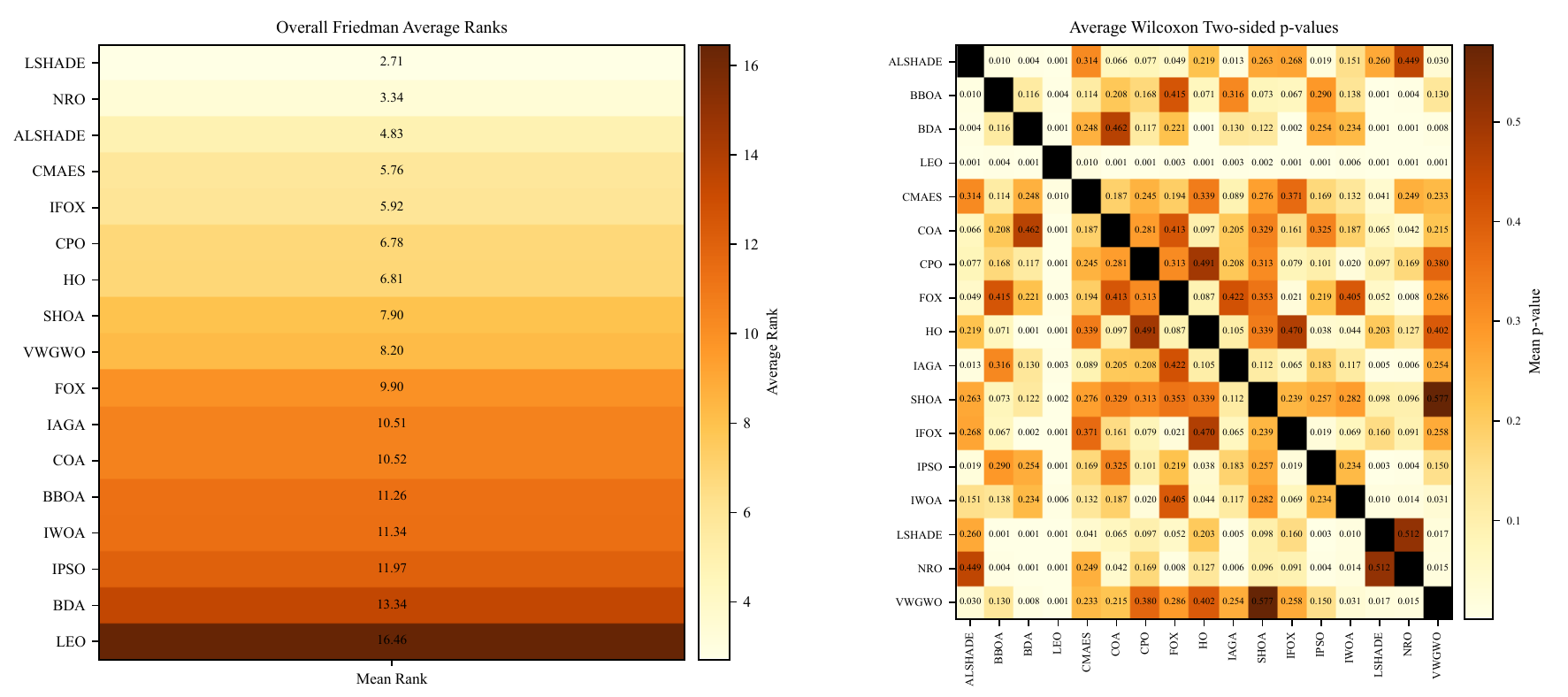}
        \caption{Average Friedman test ranks (left) and pairwise average Wilcoxon signed-rank test p-values (right) for all compared optimization algorithms across CL, C17, C19, C21, C22, and RWPs.}
        \label{fig:average_analysis_ifox}
\end{figure}

The statistical analysis from Table~\ref{tab:analysis_ifox_vs_all} showed that~\gls{ifox} was superior to~\gls{bboa},~\gls{bda},~\gls{leo},~\gls{coa},~\gls{cpo},~\gls{fox},~\gls{iaga}, and~\gls{vw-gwo} on the CL benchmark test functions. It exhibited no statistically significant difference when compared to~\gls{alshade},~\gls{cma-es},~\gls{ho},~\gls{shoa},~\gls{ipso}, and~\gls{iwoa}, but was significantly outperformed by~\gls{nro} and~\gls{lshade}. On the C17 functions,~\gls{ifox} outperformed all optimization algorithms except~\gls{alshade},~\gls{cma-es}, and~\gls{nro}, where it showed comparable performance, while it was significantly defeated by~\gls{lshade}. Moreover, for the C19 functions,~\gls{ifox} was superior to~\gls{bda},~\gls{leo},~\gls{coa},~\gls{fox},~\gls{ipso},~\gls{iwoa}, and~\gls{vw-gwo}. It performed statistically similar to~\gls{alshade},~\gls{cpo},~\gls{ho},~\gls{iaga},~\gls{shoa},~\gls{lshade}, and~\gls{nro}, but was significantly outperformed by~\gls{cma-es}. Additionally, in the C21 functions,~\gls{ifox} outperformed~\gls{bboa},~\gls{bda},~\gls{leo},~\gls{coa},~\gls{fox},~\gls{shoa},~\gls{ipso}, and~\gls{iwoa}, and showed no significant difference compared to~\gls{cpo},~\gls{ho},~\gls{iaga}, and~\gls{vw-gwo}. However, it was significantly defeated by~\gls{alshade},~\gls{cma-es},~\gls{lshade}, and~\gls{nro}. Furthermore, on C22,~\gls{ifox} demonstrated superiority over~\gls{bda},~\gls{leo},~\gls{coa},~\gls{ipso}, and~\gls{iwoa}, and showed no significant difference with~\gls{bboa},~\gls{cma-es},~\gls{fox},~\gls{ho},~\gls{iaga},~\gls{shoa}, and~\gls{vw-gwo}, while it was significantly outperformed by~\gls{alshade},~\gls{cpo},~\gls{lshade}, and~\gls{nro}. Finally, on the \glspl{rwp},~\gls{ifox} significantly outperformed~\gls{bboa},~\gls{bda},~\gls{leo},~\gls{cma-es},~\gls{fox},~\gls{iaga},~\gls{ipso}, and~\gls{iwoa}, and exhibited statistically similar performance with~\gls{alshade},~\gls{coa},~\gls{ho},~\gls{shoa},~\gls{lshade}, and~\gls{vw-gwo}, but was significantly defeated by~\gls{cpo} and~\gls{nro}. These analyses are further summarized by the pairwise comparisons in the right-hand side of Figure~\ref{fig:average_analysis_ifox}, which displays the average Wilcoxon signed-rank test p-values.  The darker cells in the heatmap reflect non-significant differences, whereas lighter cells, especially those near the 0.05 p-value, demonstrate statistically significant differences between the tested algorithms.

\begin{table}[htb]
        \caption{Wilcoxon signed-rank test results comparing IFOX with other optimization algorithms across CL, C17, C19, C21, C22, and RWPs. Values include positive ranks (R\textsuperscript{+}), negative ranks (R\textsuperscript{-}), and two-sided p-values indicating statistical significance}
        \tiny
        \begin{tabularx}{\textwidth}{|l|l|l|l|l|l|l|l|l|l|l|l|l|l|l|l|l|l|l|}
            \hline
            \multicolumn{19}{|c|}{\textbf{Wilcoxon signed-rank test of IFOX versus optimization algorithms}}                                                                                                                                                                                                                                                                                                                                                                                                                                                                                                       \\ \hline
            \textbf{Algorithm} & \multicolumn{3}{c|}{\textbf{CL}} & \multicolumn{3}{c|}{\textbf{C17}} & \multicolumn{3}{c|}{\textbf{C19}} & \multicolumn{3}{c|}{\textbf{C21}} & \multicolumn{3}{c|}{\textbf{C22}} & \multicolumn{3}{c|}{\textbf{RWPs}}                                                                                                                                                                                                                                                                                                                                                             \\ \cline{2-19}
            
                               & {\textbf{R\textsuperscript{+}}}  & {\textbf{R\textsuperscript{-}}}   & {\textbf{P-val}}                  & {\textbf{R\textsuperscript{+}}}   & {\textbf{R\textsuperscript{-}}}   & {\textbf{P-val}}                   & {\textbf{R\textsuperscript{+}}} & {\textbf{R\textsuperscript{-}}} & {\textbf{P-val}} & {\textbf{R\textsuperscript{+}}} & {\textbf{R\textsuperscript{-}}} & {\textbf{P-val}} & {\textbf{R\textsuperscript{+}}} & {\textbf{R\textsuperscript{-}}} & {\textbf{P-val}} & {\textbf{R\textsuperscript{+}}} & {\textbf{R\textsuperscript{-}}} & {\textbf{P-val}} \\ \hline 
            ALSHADE            & 59                               & 112                               & 0.25                              & 150                               & 285                               & 0.15                               & 22.0                            & 23.0                            & 1.00             & 5                               & 50                              & 0.02             & 11                              & 67                              & 0.03             & 10                              & 35                              & 0.16             \\ \hline
            BBOA               & 210                              & 0                                 & 0.00                              & 382                               & 53                                & 0.00                               & 44.0                            & 11.0                            & 0.11             & 49                              & 6                               & 0.03             & 54                              & 24                              & 0.27             & 55                              & 0                               & 0.00             \\ \hline
            BDA                & 153                              & 0                                 & 0.00                              & 435                               & 0                                 & 0.00                               & 55.0                            & 0.0                             & 0.00             & 55                              & 0                               & 0.00             & 74                              & 4                               & 0.00             & 45                              & 0                               & 0.00             \\ \hline
            LEO               & 190                              & 0                                 & 0.00                              & 435                               & 0                                 & 0.00                               & 55.0                            & 0.0                             & 0.00             & 55                              & 0                               & 0.00             & 78                              & 0                               & 0.00             & 45                              & 0                               & 0.00             \\ \hline
            CMAES              & 72                               & 64                                & 0.84                              & 200                               & 235                               & 0.72                               & 0.0                             & 45.0                            & 0.00             & 7                               & 48                              & 0.04             & 32                              & 46                              & 0.62             & 44                              & 1                               & 0.01             \\ \hline
            COA                & 75                               & 3                                 & 0.00                              & 435                               & 0                                 & 0.00                               & 45.0                            & 0.0                             & 0.00             & 55                              & 0                               & 0.00             & 65                              & 13                              & 0.04             & 21                              & 24                              & 0.91             \\ \hline
            CPO                & 111                              & 25                                & 0.03                              & 399                               & 36                                & 0.00                               & 43.0                            & 12.0                            & 0.13             & 39                              & 16                              & 0.28             & 12                              & 66                              & 0.03             & 1                               & 44                              & 0.01             \\ \hline
            FOX                & 66                               & 0                                 & 0.00                              & 429                               & 6                                 & 0.00                               & 45.0                            & 0.0                             & 0.00             & 55                              & 0                               & 0.00             & 62                              & 16                              & 0.08             & 48                              & 7                               & 0.04             \\ \hline
            HO                 & 51                               & 27                                & 0.35                              & 424                               & 11                                & 0.00                               & 24.0                            & 21.0                            & 0.91             & 39                              & 16                              & 0.28             & 29                              & 49                              & 0.47             & 20                              & 25                              & 0.82             \\ \hline
            IAGA               & 210                              & 0                                 & 0.00                              & 383                               & 52                                & 0.00                               & 42.0                            & 13.0                            & 0.16             & 47                              & 8                               & 0.05             & 57                              & 21                              & 0.18             & 55                              & 0                               & 0.00             \\ \hline
            SHOA            & 80                               & 56                                & 0.53                              & 430                               & 5                                 & 0.00                               & 43.0                            & 12.0                            & 0.13             & 50                              & 5                               & 0.02             & 45                              & 33                              & 0.68             & 7                               & 38                              & 0.07             \\ \hline
            IPSO               & 111                              & 42                                & 0.10                              & 435                               & 0                                 & 0.00                               & 55.0                            & 0.0                             & 0.00             & 55                              & 0                               & 0.00             & 78                              & 0                               & 0.00             & 44                              & 1                               & 0.01             \\ \hline
            IWOA               & 56                               & 97                                & 0.33                              & 435                               & 0                                 & 0.00                               & 55.0                            & 0.0                             & 0.00             & 48                              & 7                               & 0.04             & 66                              & 12                              & 0.03             & 44                              & 1                               & 0.01             \\ \hline
            LSHADE             & 24                               & 96                                & 0.04                              & 49                                & 386                               & 0.00                               & 24.0                            & 31.0                            & 0.77             & 5                               & 50                              & 0.02             & 0                               & 78                              & 0.00             & 9                               & 36                              & 0.13             \\ \hline
            NRO                & 19                               & 86                                & 0.04                              & 163                               & 272                               & 0.25                               & 12.0                            & 33.0                            & 0.25             & 0                               & 55                              & 0.00             & 0                               & 78                              & 0.00             & 3                               & 52                              & 0.01             \\ \hline
            VWGWO              & 120                              & 0                                 & 0.00                              & 409                               & 26                                & 0.00                               & 45.0                            & 10.0                            & 0.08             & 43                              & 12                              & 0.13             & 28                              & 50                              & 0.42             & 24                              & 21                              & 0.91             \\ \hline
        \end{tabularx}
        \label{tab:analysis_ifox_vs_all}
\end{table}

As reported in Table~\ref{tab:analysis_ifox_Friedman} and illustrated in the left-hand side of Figure~\ref{fig:average_analysis_ifox},~\gls{ifox} classified 5th overall according to the non-parametric analysis using the Friedman test, with an average rank of 5.92. This rank is ~118\% behind the~\gls{lshade} (the top algorithm), ~77\% after the~\gls{nro} (second), ~22\% behind the~\gls{alshade} (third), and ~2\% behind the~\gls{cma-es} (fourth). Achieving a rank close to these top performers---and substantially better than the original~\gls{fox}, with over 40\% improvement in ranking---indicates that the~\gls{ifox} has successfully produced an enhanced version of the original algorithm. Although the average rank of~\gls{ifox} is 118\% higher than~\gls{lshade}, it still ranks within the top five across 17 optimization algorithms that show a competitive performance. 
This improvement is primarily attributed to the proposed modifications, including the proposed parameters alpha $\alpha$ and beta $\beta$, which play a crucial role in the~\gls{ifox}. Furthermore, the modification of key~\gls{fox}'s exploration and exploitation equations, Eq~(\ref{eq:x_exploit_ifox}) and~(\ref{eq:x_explore_ifox}). In addition to the replacing of static balance method of~\gls{fox} with the proposed adaptive method in Algorithm~\ref{algo:ifox_balance}. These modifications collectively contribute to producing a competitive optimization algorithm. Regardless, the~\gls{ifox} still has limitations, which are discussed in the following section.

\begin{table}[htb]
    \caption{Friedman test average rankings of IFOX and competing optimization algorithms across CL, C17, C19, C21, C22, and RWPs. The "Position" column indicates the overall rank based on the average performance.}
    \label{tab:analysis_ifox_Friedman}
    \begin{tabularx}{\textwidth}{|X|X|X|X|X|X|X|X|X|}
        \hline
        \multicolumn{ 1}{|l|}{\textbf{Algorithm}} & \multicolumn{ 6}{c|}{\textbf{Friedman test ranking/Functions}} & \multicolumn{ 1}{l|}{\textbf{Mean}} & \multicolumn{ 1}{l|}{\textbf{Position}}                                                                                                   \\ \cline{ 2- 7}
        \multicolumn{ 1}{|l|}{}                   & \textbf{CL}                                                    & \textbf{C17}                        & \textbf{C19}                            & \textbf{C21}  & \textbf{C22}  & \textbf{RWPs} & \multicolumn{ 1}{l|}{} & \multicolumn{ 1}{l|}{} \\ \hline
        LSHADE                                    & 2.75                                                           & 1.90                                & 3.70                                    & 3.30          & 1.92          & 2.70          & 2.71                   & 1                      \\ \hline
        NRO                                       & 2.55                                                           & 4.14                                & 3.50                                    & 2.80          & 2.25          & 4.80          & 3.34                   & 2                      \\ \hline
        ALSHADE                                   & 7.00                                                           & 4.45                                & 6.20                                    & 3.40          & 4.25          & 3.70          & 4.83                   & 3                      \\ \hline
        CMAES                                     & 6.75                                                           & 4.48                                & 1.30                                    & 2.70          & 6.42          & 12.90         & 5.76                   & 4                      \\ \hline
        \textbf{IFOX}                             & \textbf{4.65}                                                  & \textbf{4.72}                       & \textbf{5.30}                           & \textbf{6.50} & \textbf{7.83} & \textbf{6.50} & \textbf{5.92 }         & \textbf{5}             \\ \hline
        CPO                                       & 7.55                                                           & 7.38                                & 8.40                                    & 6.90          & 5.67          & 4.80          & 6.78                   & 6                      \\ \hline
        HO                                        & 4.55                                                           & 9.45                                & 5.90                                    & 8.00          & 6.83          & 6.10          & 6.81                   & 7                      \\ \hline
        SHOA                                   & 5.70                                                           & 11.79                               & 9.90                                    & 8.40          & 7.50          & 4.10          & 7.90                   & 8                      \\ \hline
        VWGWO                                     & 8.00                                                           & 8.76                                & 10.20                                   & 8.00          & 7.92          & 6.30          & 8.20                   & 9                      \\ \hline
        FOX                                       & 6.75                                                           & 7.72                                & 10.70                                   & 11.10         & 12.25         & 10.90         & 9.90                   & 10                     \\ \hline
        IAGA                                      & 14.55                                                          & 7.38                                & 8.80                                    & 9.70          & 10.33         & 12.30         & 10.51                  & 11                     \\ \hline
        COA                                       & 5.95                                                           & 14.41                               & 11.30                                   & 13.40         & 11.75         & 6.30          & 10.52                  & 12                     \\ \hline
        BBOA                                      & 15.65                                                          & 7.45                                & 10.20                                   & 11.00         & 10.67         & 12.60         & 11.26                  & 13                     \\ \hline
        IWOA                                      & 4.80                                                           & 13.41                               & 12.40                                   & 12.30         & 13.75         & 11.40         & 11.34                  & 14                     \\ \hline
        IPSO                                      & 7.00                                                           & 13.90                               & 12.80                                   & 13.90         & 13.92         & 10.30         & 11.97                  & 15                     \\ \hline
        BDA                                       & 11.05                                                          & 14.69                               & 13.10                                   & 14.70         & 12.50         & 14.00         & 13.34                  & 16                     \\ \hline
        LEO                                      & 15.80                                                          & 16.97                               & 16.90                                   & 16.90         & 16.92         & 15.30         & 16.46                  & 17                     \\ \hline
    \end{tabularx}
\end{table}

\subsection{Limitations and scalability analysis}
Identifying the limitations of the proposed~\gls{ifox} was crucial to pinpoint its strengths and weaknesses. To this end, we selected test function C17F8---on which neither~\gls{ifox} nor competing optimizers reached the optimal objective value 800---and ran it at four problem dimensions (10, 30, 50, 100) and five epoch settings (50, 100, 250, 500, 1000). The results have been averaged over 30 independent runs as presented in Table~\ref{tab:scalability_results}.

\begin{figure}[htb]
        \centering
        \includegraphics[width=1\textwidth]{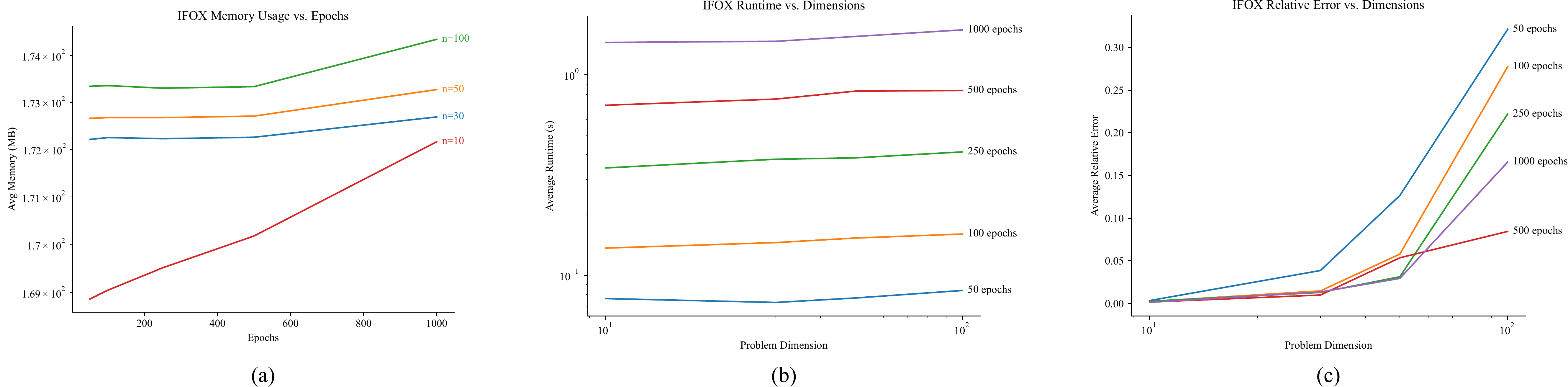}
        \caption{Scalability of IFOX on C17F8: (a) average memory usage, (b) average runtime, and (c) average relative error across epochs (50–1000) for dimensions 10, 30, 50, and 100.}
        \label{fig:scalability_analysis_ifox}
\end{figure}

As shown in Figure~\ref{fig:scalability_analysis_ifox}(a), memory consumption increases only slightly with both epochs and dimension: at dimension 10, memory usage rises from 168.9 MB at 50 epochs to 172.2 MB at 1000 epochs (1.9\%), while at 1000 epochs scaling from dimension 10 to 100 yields an additional 2.2 MB (1.3\%). On the other hand, runtime scales nearly linearly with epochs as shown in Figure~\ref{fig:scalability_analysis_ifox}(b): at dimension 10, average runtime grows from 0.30 s to 6.19 s (20×) when increasing epochs from 50 to 1000; similarly, at 1000 epochs, increasing the dimension from 10 to 100 raises runtime by 17\% (6.19 s to 7.23 s). Lastly, the relative error remains very low for low-dimensional cases, decreasing from 0.00233 at 50 epochs to 0.00174 at 500 epochs for dimension 10, but exhibits significant degradation at higher dimensions as shown in Figure~\ref{fig:scalability_analysis_ifox}(c). For dimension 100, relative error peaks at 0.1936 for 50 epochs and plateaus near 0.1178 even at 1000 epochs, which indicates slower convergence.

The scalability analysis demonstrates that IFOX keeps the same trends for memory use and running duration when facing increases in problem dimensions which demonstrates the notable advantages of this algorithm. IFOX faces a major limitation due to its increased relative error in finding the optimal objective value between low and high-dimensional problems. Thus, future investigation needs to develop methods for adaptive epoch control according to problem dimensions and elastic search strategies to boost convergence results in big-dimensional problems. Furthermore, utilizing techniques such as quantum computing and parallel processing with IFOX can potentially lead to improved overall performance.

\begin{table}[htb]
    \caption{Scalability of the proposed IFOX on C17F8 with the following metrics: Runtime (seconds), Memory usage (MB), and Relative error across problem dimensions (10, 30, 50, 100) and epochs (50, 100, 250, 500, 1000), averaged over 30 independent runs.}
    \label{tab:scalability_results}
    \begin{tabularx}{\textwidth}{|l|l|X|X|X|}
        \hline
        \textbf{Dim}                        & \textbf{Epoch} & \textbf{Runtime (seconds)} & \textbf{Memory usage (MB)} & \textbf{Relative error} \\ \hline
        \multicolumn{ 1}{|l|}{\textbf{10}}  & 50             & 0.304                      & 168.865                    & 0.002                   \\ \cline{ 2- 5}
        \multicolumn{ 1}{|l|}{}             & 100            & 0.611                      & 169.051                    & 0.002                   \\ \cline{ 2- 5}
        \multicolumn{ 1}{|l|}{}             & 250            & 1.549                      & 169.517                    & 0.002                   \\ \cline{ 2- 5}
        \multicolumn{ 1}{|l|}{}             & 500            & 3.065                      & 170.181                    & 0.002                   \\ \cline{ 2- 5}
        \multicolumn{ 1}{|l|}{}             & 1000           & 6.185                      & 172.165                    & 0.002                   \\ \hline
        \multicolumn{ 1}{|l|}{\textbf{30}}  & 50             & 0.325                      & 172.214                    & 0.019                   \\ \cline{ 2- 5}
        \multicolumn{ 1}{|l|}{}             & 100            & 0.642                      & 172.254                    & 0.014                   \\ \cline{ 2- 5}
        \multicolumn{ 1}{|l|}{}             & 250            & 1.599                      & 172.230                    & 0.012                   \\ \cline{ 2- 5}
        \multicolumn{ 1}{|l|}{}             & 500            & 3.234                      & 172.261                    & 0.014                   \\ \cline{ 2- 5}
        \multicolumn{ 1}{|l|}{}             & 1000           & 6.360                      & 172.692                    & 0.011                   \\ \hline
        \multicolumn{ 1}{|l|}{\textbf{50}}  & 50             & 0.330                      & 172.663                    & 0.063                   \\ \cline{ 2- 5}
        \multicolumn{ 1}{|l|}{}             & 100            & 0.657                      & 172.677                    & 0.043                   \\ \cline{ 2- 5}
        \multicolumn{ 1}{|l|}{}             & 250            & 1.642                      & 172.677                    & 0.043                   \\ \cline{ 2- 5}
        \multicolumn{ 1}{|l|}{}             & 500            & 3.294                      & 172.709                    & 0.038                   \\ \cline{ 2- 5}
        \multicolumn{ 1}{|l|}{}             & 1000           & 6.543                      & 173.274                    & 0.039                   \\ \hline
        \multicolumn{ 1}{|l|}{\textbf{100}} & 50             & 0.370                      & 173.344                    & 0.194                   \\ \cline{ 2- 5}
        \multicolumn{ 1}{|l|}{}             & 100            & 0.727                      & 173.358                    & 0.133                   \\ \cline{ 2- 5}
        \multicolumn{ 1}{|l|}{}             & 250            & 1.814                      & 173.304                    & 0.110                   \\ \cline{ 2- 5}
        \multicolumn{ 1}{|l|}{}             & 500            & 3.656                      & 173.337                    & 0.116                   \\ \cline{ 2- 5}
        \multicolumn{ 1}{|l|}{}             & 1000           & 7.227                      & 174.348                    & 0.118                   \\ \hline
    \end{tabularx}
\end{table}

\section{Conclusion\label{sec:conclusion}}
In conclusion, this paper introduces~\gls{ifox} as a new improved optimization algorithm which surpasses~\gls{fox} through its multiple improvements. Unlike the static balance in~\gls{fox},~\gls{ifox} suggests an adaptive method to balance exploration and exploitation dynamically. It also reduces the number of hyperparameters and modifies the primary equations to enhance performance and address existing limitations. The performance of~\gls{ifox} has been evaluated using 81 benchmark test functions and ten real-world problems. The results show that~\gls{ifox} has 880 wins, 228 ties, and 348 losses against the established optimization algorithms across all tested functions and problems. This substantial improvement highlights the effectiveness of the~\gls{ifox} in solving diverse optimization tasks and real-world problems. However,~\gls{ifox} also presents certain limitations. Scalability analysis revealed that while memory usage and runtime increase moderately with higher dimensions and epochs, the relative error in high-dimensional problems remains a challenge, indicating a slower convergence rate. Future work should focus on incorporating adaptive epoch control mechanisms based on problem dimensionality and integrating elastic search strategies to mitigate performance degradation in high-dimensional spaces. Moreover, leveraging advanced computational techniques such as parallel processing or quantum computing may further enhance the scalability and applicability of~\gls{ifox} to real-time and large-scale optimization problems.

\bibliographystyle{elsarticle-num}

\begin{thebibliography}{10}

\bibitem{optimizatoin_1}
Rahimi I, Gandomi AH, Chen F, Mezura-Montes E.
\newblock A Review on Constraint Handling Techniques for Population-based Algorithms: from single-objective to multi-objective optimization.
\newblock Archives of Computational Methods in Engineering. 2022;30(3):2181–2209.
\newblock doi:{10.1007/s11831-022-09859-9}.

\bibitem{life_problems_1}
He C, Zhang Y, Gong D, Ji X.
\newblock A review of surrogate-assisted evolutionary algorithms for expensive optimization problems.
\newblock Expert Systems with Applications. 2023;217:119495.
\newblock doi:{https://doi.org/10.1016/j.eswa.2022.119495}.

\bibitem{benefits_1}
ZHU J, ZHOU H, WANG C, ZHOU L, YUAN S, ZHANG W.
\newblock A review of topology optimization for additive manufacturing: Status and challenges.
\newblock Chinese Journal of Aeronautics. 2021;34(1):91--110.
\newblock doi:{https://doi.org/10.1016/j.cja.2020.09.020}.

\bibitem{industry_1}
Pinciroli L, Baraldi P, Zio E.
\newblock Maintenance optimization in industry 4.0.
\newblock Reliability Engineering \&; System Safety. 2023;234:109204.
\newblock doi:{10.1016/j.ress.2023.109204}.

\bibitem{uot_2}
Mutar H, Jawad M.
\newblock Analytical Study for Optimization Techniques to Prolong WSNs Life.
\newblock IRAQI JOURNAL OF COMPUTERS, COMMUNICATIONS, CONTROL AND SYSTEMS ENGINEERING. 2023;23(2):13--23.
\newblock doi:{https://doi.org/10.33103/uot.ijccce.23.2.2}.

\bibitem{optimizer_1}
Sadatdiynov K, Cui L, Zhang L, Huang JZ, Salloum S, Mahmud MS.
\newblock A review of optimization methods for computation offloading in edge computing networks.
\newblock Digital Communications and Networks. 2023;9(2):450--461.
\newblock doi:{https://doi.org/10.1016/j.dcan.2022.03.003}.

\bibitem{benefits_2}
Osaba E, Villar-Rodriguez E, {Del Ser} J, Nebro AJ, Molina D, LaTorre A, et~al.
\newblock A Tutorial On the design, experimentation and application of metaheuristic algorithms to real-World optimization problems.
\newblock Swarm and Evolutionary Computation. 2021;64:100888.
\newblock doi:{https://doi.org/10.1016/j.swevo.2021.100888}.

\bibitem{optimizer_types_1}
Hauswirth A, He Z, Bolognani S, Hug G, Dörfler F.
\newblock Optimization algorithms as robust feedback controllers.
\newblock Annual Reviews in Control. 2024;57:100941.
\newblock doi:{https://doi.org/10.1016/j.arcontrol.2024.100941}.

\bibitem{life_problems_2}
Zhu F, Li G, Tang H, Li Y, Lv X, Wang X.
\newblock Dung beetle optimization algorithm based on quantum computing and multi-strategy fusion for solving engineering problems.
\newblock Expert Systems with Applications. 2024;236:121219.
\newblock doi:{10.1016/j.eswa.2023.121219}.

\bibitem{fox}
Mohammed H, Rashid T.
\newblock FOX: a FOX-inspired optimization algorithm.
\newblock Applied Intelligence. 2022;53(1):1030–1050.
\newblock doi:{10.1007/s10489-022-03533-0}.

\bibitem{rastrigin_figure}
Sch\"{o}nenberger L, Beyer HG.
\newblock On a Population Sizing Model for Evolution Strategies Optimizing the Highly Multimodal Rastrigin Function.
\newblock In: Proceedings of the Genetic and Evolutionary Computation Conference. GECCO ’23. ACM; 2023.Available from: \url{http://dx.doi.org/10.1145/3583131.3590451}.

\bibitem{related_optimizatoin_in_cs}
Choi H, Park S.
\newblock A Survey of Machine Learning-Based System Performance Optimization Techniques.
\newblock Applied Sciences. 2021;11(7).
\newblock doi:{10.3390/app11073235}.

\bibitem{related_optimizatoin_problems_1}
Salawudeen AT, Mu’azu MB, Sha’aban YA, Adedokun AE.
\newblock A Novel Smell Agent Optimization (SAO): An extensive CEC study and engineering application.
\newblock Knowledge-Based Systems. 2021;232:107486.
\newblock doi:{https://doi.org/10.1016/j.knosys.2021.107486}.

\bibitem{related_optimizatoin_problems_2}
Vagaska A, Gombar M, Straka L.
\newblock Selected Mathematical Optimization Methods for Solving Problems of Engineering Practice.
\newblock Energies. 2022;15(6).
\newblock doi:{10.3390/en15062205}.

\bibitem{related_optimizatoin_problems_3}
Tang J, Liu G, Pan Q.
\newblock A Review on Representative Swarm Intelligence Algorithms for Solving Optimization Problems: Applications and Trends.
\newblock IEEE/CAA Journal of Automatica Sinica. 2021;8(10):1627--1643.
\newblock doi:{10.1109/JAS.2021.1004129}.

\bibitem{related_hot_topic_area}
Pardalos PM, Romeijn HE, Tuy H.
\newblock Recent developments and trends in global optimization.
\newblock Journal of Computational and Applied Mathematics. 2000;124(1):209--228.
\newblock doi:{https://doi.org/10.1016/S0377-0427(00)00425-8}.

\bibitem{uot_1}
Jabber S, Hashem S, Jafer S.
\newblock Analytical and Comparative Study for Optimization Problems.
\newblock IRAQI JOURNAL OF COMPUTERS, COMMUNICATIONS, CONTROL AND SYSTEMS ENGINEERING. 2023;23(4):46--57.
\newblock doi:{https://doi.org/10.33103/uot.ijccce.23.4.5}.

\bibitem{swarm_intelligence}
Tang J, Liu G, Pan Q.
\newblock A Review on Representative Swarm Intelligence Algorithms for Solving Optimization Problems: Applications and Trends.
\newblock IEEE/CAA Journal of Automatica Sinica. 2021;8(10):1627–1643.
\newblock doi:{10.1109/jas.2021.1004129}.

\bibitem{swarm_intelligence1}
Tang J, Duan H, Lao S.
\newblock Swarm intelligence algorithms for multiple unmanned aerial vehicles collaboration: a comprehensive review.
\newblock Artificial Intelligence Review. 2022;56(5):4295–4327.
\newblock doi:{10.1007/s10462-022-10281-7}.

\bibitem{related_SHOA}
AbdulKarim HK, Rashid TA.
\newblock In Search of Excellence: SHOA as a Competitive Shrike Optimization Algorithm for Multimodal Problems.
\newblock IEEE Access. 2024;12:98407–98425.
\newblock doi:{10.1109/access.2024.3427632}.

\bibitem{improved_gwo_velocity_based}
Rezaei F, Safavi HR, Abd~Elaziz M, El-Sappagh SHA, Al-Betar MA, Abuhmed T.
\newblock An Enhanced Grey Wolf Optimizer with a Velocity-Aided Global Search Mechanism.
\newblock Mathematics. 2022;10(3):351.
\newblock doi:{10.3390/math10030351}.

\bibitem{enhanced_Search_gwo}
Shial G, Sahoo S, Panigrahi S.
\newblock An Enhanced GWO Algorithm with Improved Explorative Search Capability for Global Optimization and Data Clustering.
\newblock Applied Artificial Intelligence. 2023;37(1).
\newblock doi:{10.1080/08839514.2023.2166232}.

\bibitem{gwo_fs}
Pan H, Chen S, Xiong H.
\newblock A high-dimensional feature selection method based on modified Gray Wolf Optimization.
\newblock Applied Soft Computing. 2023;135:110031.
\newblock doi:{10.1016/j.asoc.2023.110031}.

\bibitem{related_vw-gwo}
Gao ZM, Zhao J.
\newblock An Improved Grey Wolf Optimization Algorithm with Variable Weights.
\newblock Computational Intelligence and Neuroscience. 2019;2019:1–13.
\newblock doi:{10.1155/2019/2981282}.

\bibitem{related_power_optimization}
Li X, Wang Z, Yang C, Bozkurt A.
\newblock An advanced framework for net electricity consumption prediction: Incorporating novel machine learning models and optimization algorithms.
\newblock Energy. 2024;296:131259.
\newblock doi:{10.1016/j.energy.2024.131259}.

\bibitem{related_bboa}
Prakash T, Singh PP, Singh VP, Singh SN.
\newblock A novel Brown-bear optimization algorithm for solving economic dispatch problem.
\newblock In: Advanced Control and Optimization Paradigms for Energy System Operation and Management. River Publishers; 2023. p. 137--164.
\newblock Available from: \url{http://dx.doi.org/10.1201/9781003337003}.

\bibitem{related_da}
Mirjalili S.
\newblock Dragonfly algorithm: a new meta-heuristic optimization technique for solving single-objective, discrete, and multi-objective problems.
\newblock Neural Computing and Applications. 2015;27(4):1053–1073.
\newblock doi:{10.1007/s00521-015-1920-1}.

\bibitem{related_bda}
Hammouri AI, Mafarja M, Al-Betar MA, Awadallah MA, Abu-Doush I.
\newblock An improved Dragonfly Algorithm for feature selection.
\newblock Knowledge-Based Systems. 2020;203:106131.
\newblock doi:{10.1016/j.knosys.2020.106131}.

\bibitem{related_bat}
Yang XS, Hossein~Gandomi A.
\newblock Bat algorithm: a novel approach for global engineering optimization.
\newblock Engineering computations. 2012;29(5):464--483.
\newblock doi:{10.1108/02644401211235834}.

\bibitem{related_chba}
Yu H, Zhao N, Wang P, Chen H, Li C.
\newblock Chaos-enhanced synchronized bat optimizer.
\newblock Applied Mathematical Modelling. 2020;77:1201–1215.
\newblock doi:{10.1016/j.apm.2019.09.029}.

\bibitem{related_ipso}
Qiuyun T, Hongyan S, Hengwei G, Ping W.
\newblock Improved Particle Swarm Optimization Algorithm for AGV Path Planning.
\newblock IEEE Access. 2021;9:33522–33531.
\newblock doi:{10.1109/access.2021.3061288}.

\bibitem{related_iaga}
Han S, Xiao L.
\newblock An improved adaptive genetic algorithm.
\newblock SHS Web of Conferences. 2022;140:01044.
\newblock doi:{10.1051/shsconf/202214001044}.

\bibitem{related_iwoa}
Mostafa~Bozorgi S, Yazdani S.
\newblock IWOA: An improved whale optimization algorithm for optimization problems.
\newblock Journal of Computational Design and Engineering. 2019;6(3):243–259.
\newblock doi:{10.1016/j.jcde.2019.02.002}.

\bibitem{lshade}
Piotrowski AP.
\newblock L-SHADE optimization algorithms with population-wide inertia.
\newblock Information Sciences. 2018;468:117–141.
\newblock doi:{10.1016/j.ins.2018.08.030}.

\bibitem{alshade}
Li Y, Han T, Zhou H, Tang S, Zhao H.
\newblock A novel adaptive L-SHADE algorithm and its application in UAV swarm resource configuration problem.
\newblock Information Sciences. 2022;606:350–367.
\newblock doi:{10.1016/j.ins.2022.05.058}.

\bibitem{cma-es}
Hansen N, Müller SD, Koumoutsakos P.
\newblock Reducing the Time Complexity of the Derandomized Evolution Strategy with Covariance Matrix Adaptation (CMA-ES).
\newblock Evolutionary Computation. 2003;11(1):1--18.
\newblock doi:{10.1162/106365603321828970}.

\bibitem{cpo}
Abdel-Basset M, Mohamed R, Abouhawwash M.
\newblock Crested Porcupine Optimizer: A new nature-inspired metaheuristic.
\newblock Knowledge-Based Systems. 2024;284:111257.
\newblock doi:{10.1016/j.knosys.2023.111257}.

\bibitem{coa}
Dehghani M, Montazeri Z, Trojovská E, Trojovský P.
\newblock Coati Optimization Algorithm: A new bio-inspired metaheuristic algorithm for solving optimization problems.
\newblock Knowledge-Based Systems. 2023;259:110011.
\newblock doi:{10.1016/j.knosys.2022.110011}.

\bibitem{HO}
Amiri MH, Mehrabi~Hashjin N, Montazeri M, Mirjalili S, Khodadadi N.
\newblock Hippopotamus optimization algorithm: a novel nature-inspired optimization algorithm.
\newblock Scientific Reports. 2024;14(1).
\newblock doi:{10.1038/s41598-024-54910-3}.

\bibitem{related_nro}
Wei Z, Huang C, Wang X, Han T, Li Y.
\newblock Nuclear Reaction Optimization: A Novel and Powerful Physics-Based Algorithm for Global Optimization.
\newblock IEEE Access. 2019;7:66084–66109.
\newblock doi:{10.1109/access.2019.2918406}.

\bibitem{leo2023}
Aladdin AM, Rashid TA.
\newblock LEO: Lagrange elementary optimization.
\newblock Neural Computing and Applications. 2025;doi:{10.1007/s00521-025-11225-2}.

\bibitem{ao_optimizer}
Abualigah L, Yousri D, Abd~Elaziz M, Ewees AA, Al-qaness MAA, Gandomi AH.
\newblock Aquila Optimizer: A novel meta-heuristic optimization algorithm.
\newblock Computers \&amp; Industrial Engineering. 2021;157:107250.
\newblock doi:{10.1016/j.cie.2021.107250}.

\bibitem{rsa_optimizer}
Abualigah L, Elaziz MA, Sumari P, Geem ZW, Gandomi AH.
\newblock Reptile Search Algorithm (RSA): A nature-inspired meta-heuristic optimizer.
\newblock Expert Systems with Applications. 2022;191:116158.
\newblock doi:{10.1016/j.eswa.2021.116158}.

\bibitem{gto_optimizer}
Abdollahzadeh B, Soleimanian~Gharehchopogh F, Mirjalili S.
\newblock Artificial gorilla troops optimizer: A new nature‐inspired metaheuristic algorithm for global optimization problems.
\newblock International Journal of Intelligent Systems. 2021;36(10):5887–5958.
\newblock doi:{10.1002/int.22535}.

\bibitem{sma_optimizer}
Li S, Chen H, Wang M, Heidari AA, Mirjalili S.
\newblock Slime mould algorithm: A new method for stochastic optimization.
\newblock Future Generation Computer Systems. 2020;111:300–323.
\newblock doi:{10.1016/j.future.2020.03.055}.

\bibitem{bmo_optimizer}
Sulaiman MH, Mustaffa Z, Saari MM, Daniyal H.
\newblock Barnacles Mating Optimizer: A new bio-inspired algorithm for solving engineering optimization problems.
\newblock Engineering Applications of Artificial Intelligence. 2020;87:103330.
\newblock doi:{10.1016/j.engappai.2019.103330}.

\bibitem{bssa_optimizer}
Ghasemi-Marzbali A.
\newblock A novel nature-inspired meta-heuristic algorithm for optimization: bear smell search algorithm.
\newblock Soft Computing. 2020;24(17):13003–13035.
\newblock doi:{10.1007/s00500-020-04721-1}.

\bibitem{bwoa_optimizer}
Hayyolalam V, Pourhaji~Kazem AA.
\newblock Black Widow Optimization Algorithm: A novel meta-heuristic approach for solving engineering optimization problems.
\newblock Engineering Applications of Artificial Intelligence. 2020;87:103249.
\newblock doi:{10.1016/j.engappai.2019.103249}.

\bibitem{mrfo_optimizer}
Zhao W, Zhang Z, Wang L.
\newblock Manta ray foraging optimization: An effective bio-inspired optimizer for engineering applications.
\newblock Engineering Applications of Artificial Intelligence. 2020;87:103300.
\newblock doi:{10.1016/j.engappai.2019.103300}.

\bibitem{mpa_optimizer}
Faramarzi A, Heidarinejad M, Mirjalili S, Gandomi AH.
\newblock Marine Predators Algorithm: A nature-inspired metaheuristic.
\newblock Expert Systems with Applications. 2020;152:113377.
\newblock doi:{10.1016/j.eswa.2020.113377}.

\bibitem{ma_optimizer}
Zervoudakis K, Tsafarakis S.
\newblock A mayfly optimization algorithm.
\newblock Computers \&amp; Industrial Engineering. 2020;145:106559.
\newblock doi:{10.1016/j.cie.2020.106559}.

\bibitem{survey_optimization_2023}
Alorf A.
\newblock A survey of recently developed metaheuristics and their comparative analysis.
\newblock Engineering Applications of Artificial Intelligence. 2023;117:105622.
\newblock doi:{10.1016/j.engappai.2022.105622}.

\bibitem{survey_optimization_2023_1}
Toaza B, Esztergár-Kiss D.
\newblock A review of metaheuristic algorithms for solving TSP-based scheduling optimization problems.
\newblock Applied Soft Computing. 2023;148:110908.
\newblock doi:{10.1016/j.asoc.2023.110908}.

\bibitem{survey_optimization_2025}
Sadana U, Chenreddy A, Delage E, Forel A, Frejinger E, Vidal T.
\newblock A survey of contextual optimization methods for decision-making under uncertainty.
\newblock European Journal of Operational Research. 2025;320(2):271–289.
\newblock doi:{10.1016/j.ejor.2024.03.020}.

\bibitem{related_aco}
Dorigo M, Birattari M, Stutzle T.
\newblock Ant colony optimization.
\newblock IEEE Computational Intelligence Magazine. 2006;1(4):28--39.
\newblock doi:{10.1109/MCI.2006.329691}.

\bibitem{why_needs_test_optimizer}
Osaba E, Villar-Rodriguez E, Del~Ser J, Nebro AJ, Molina D, LaTorre A, et~al.
\newblock A Tutorial On the design, experimentation and application of metaheuristic algorithms to real-World optimization problems.
\newblock Swarm and Evolutionary Computation. 2021;64:100888.
\newblock doi:{10.1016/j.swevo.2021.100888}.

\bibitem{sphare_test_function}
Garzón M, Álvarez Pomar L, Rojas-Galeano S.
\newblock From collective intelligence to global optimisation: an agent-based model approach.
\newblock Computing. 2025;107(3).
\newblock doi:{10.1007/s00607-025-01429-8}.

\bibitem{classical_optimization_functions}
Jamil M, Yang XS.
\newblock A literature survey of benchmark functions for global optimisation problems.
\newblock International Journal of Mathematical Modelling and Numerical Optimisation. 2013;4(2):150.
\newblock doi:{10.1504/ijmmno.2013.055204}.

\bibitem{benchmark_test_functions}
Sharma P, Raju S.
\newblock Metaheuristic optimization algorithms: a comprehensive overview and classification of benchmark test functions.
\newblock Soft Computing. 2023;28(4):3123–3186.
\newblock doi:{10.1007/s00500-023-09276-5}.

\bibitem{cec2017}
Škvorc U, Eftimov T, Korošec P.
\newblock CEC Real-Parameter Optimization Competitions: Progress from 2013 to 2018.
\newblock In: 2019 IEEE Congress on Evolutionary Computation (CEC); 2019. p. 3126--3133.

\bibitem{cec2019}
Brest J, Maučec MS, Bošković B.
\newblock The 100-Digit Challenge: Algorithm jDE100.
\newblock In: 2019 IEEE Congress on Evolutionary Computation (CEC); 2019. p. 19--26.
\newblock Available from: \url{https://doi.org/10.1109/CEC.2019.8789904}.

\bibitem{cec2021}
Bujok P, Kolenovsky P.
\newblock Differential Evolution with Distance-based Mutation-selection Applied to CEC 2021 Single Objective Numerical Optimisation.
\newblock In: 2021 IEEE Congress on Evolutionary Computation (CEC); 2021. p. 849--856.
\newblock Available from: \url{https://doi.org/10.1109/CEC45853.2021.9504795}.

\bibitem{cec2022}
Bujok P, Kolenovsky P.
\newblock Eigen Crossover in Cooperative Model of Evolutionary Algorithms Applied to CEC 2022 Single Objective Numerical Optimisation.
\newblock In: 2022 IEEE Congress on Evolutionary Computation (CEC); 2022. p. 1--8.
\newblock Available from: \url{https://doi.org/10.1109/CEC55065.2022.9870433}.

\bibitem{pdo_2022}
Ezugwu AE, Agushaka JO, Abualigah L, Mirjalili S, Gandomi AH.
\newblock Prairie Dog Optimization Algorithm.
\newblock Neural Computing and Applications. 2022;34(22):20017–20065.
\newblock doi:{10.1007/s00521-022-07530-9}.

\bibitem{ihaoavoa_2022}
Xiao Y, Guo Y, Cui H, Wang Y, Li J, Zhang Y.
\newblock IHAOAVOA: An improved hybrid aquila optimizer and African vultures optimization algorithm for global optimization problems.
\newblock Mathematical Biosciences and Engineering. 2022;19(11):10963–11017.
\newblock doi:{10.3934/mbe.2022512}.

\bibitem{moeosma_2023}
Luo Q, Yin S, Zhou G, Meng W, Zhao Y, Zhou Y.
\newblock Multi-objective equilibrium optimizer slime mould algorithm and its application in solving engineering problems.
\newblock Structural and Multidisciplinary Optimization. 2023;66(5).
\newblock doi:{10.1007/s00158-023-03568-y}.

\bibitem{pvd}
Moyya S, Thejasree P, Cherian~Abraham B, Mangalathu GS.
\newblock Design and analysis of single and multi-layer pressure vessel.
\newblock Materials Today: Proceedings. 2023;doi:{10.1016/j.matpr.2023.06.393}.

\bibitem{pvd_mathmatical}
Moyya S, Thejasree P, Cherian~Abraham B, Mangalathu GS.
\newblock Design and analysis of single and multi-layer pressure vessel.
\newblock Materials Today: Proceedings. 2023;doi:{10.1016/j.matpr.2023.06.393}.

\bibitem{foxann}
Jumaah MA, Ali YH, Rashid TA, Vimal S.
\newblock FOXANN: A Method for Boosting Neural Network Performance.
\newblock Journal of Soft Computing and Computer Applications. 2024;1(1).
\newblock doi:{10.70403/3008-1084.1001}.

\bibitem{fox_tsa}
Aula SA, Rashid TA.
\newblock FOX-TSA: Navigating Complex Search Spaces and Superior Performance in Benchmark and Real-World Optimization Problems.
\newblock Ain Shams Engineering Journal. 2025;16(1):103185.
\newblock doi:{10.1016/j.asej.2024.103185}.

\bibitem{used_fox_1}
ALRahhal H, Jamous R.
\newblock AFOX: a new adaptive nature-inspired optimization algorithm.
\newblock Artificial Intelligence Review. 2023;56(12):15523–15566.
\newblock doi:{10.1007/s10462-023-10542-z}.

\bibitem{used_fox_2}
ALRahhal H, Jamous R.
\newblock RNN-AFOX: adaptive FOX-inspired-based technique for automated tuning of recurrent neural network hyper-parameters.
\newblock Artificial Intelligence Review. 2023;56(S2):1981–2011.
\newblock doi:{10.1007/s10462-023-10568-3}.

\bibitem{used_fox_3}
Feda AK, Adegboye M, Adegboye OR, Agyekum EB, Fendzi~Mbasso W, Kamel S.
\newblock S-shaped grey wolf optimizer-based FOX algorithm for feature selection.
\newblock Heliyon. 2024;10(2):e24192--e24192.
\newblock doi:{10.1016/j.heliyon.2024.e24192}.

\bibitem{soundSpeed}
Sauerheber R.
\newblock Light and Sound Speed Mechanisms, Relative Velocities, Simultaneity, and Special Relativity.
\newblock SSRN Electronic Journal. 2024;doi:{10.2139/ssrn.4780574}.

\bibitem{optimizationYossra}
Hussain~Ali Y, Sabu~Chooralil V, Balasubramanian K, Manyam RR, Kidambi~Raju S, T~Sadiq A, et~al.
\newblock Optimization System Based on Convolutional Neural Network and Internet of Medical Things for Early Diagnosis of Lung Cancer.
\newblock Bioengineering. 2023;10(3):320.
\newblock doi:{10.3390/bioengineering10030320}.

\bibitem{levy}
Kaidi W, Khishe M, Mohammadi M.
\newblock Dynamic Levy Flight Chimp Optimization.
\newblock Knowledge-Based Systems. 2022;235:107625.
\newblock doi:{10.1016/j.knosys.2021.107625}.

\bibitem{fdo}
Abdullah JM, Ahmed T.
\newblock Fitness Dependent Optimizer: Inspired by the Bee Swarming Reproductive Process.
\newblock IEEE Access. 2019;7:43473--43486.
\newblock doi:{10.1109/ACCESS.2019.2907012}.

\bibitem{friedman_test}
Liu J, Xu Y.
\newblock T-Friedman Test: A New Statistical Test for Multiple Comparison with an Adjustable Conservativeness Measure.
\newblock International Journal of Computational Intelligence Systems. 2022;15(1).
\newblock doi:{10.1007/s44196-022-00083-8}.

\bibitem{statistical_analysis}
Wagenmakers EJ, Sarafoglou A, Aczel B.
\newblock One statistical analysis must not rule them all.
\newblock Nature. 2022;605(7910):423–425.
\newblock doi:{10.1038/d41586-022-01332-8}.

\end{thebibliography}

\end{document}